\documentclass{article}

\usepackage{iclr2027_conference,times}

\usepackage{amsmath}
\usepackage{amsfonts}
\usepackage{amssymb}
\usepackage{bm}
\usepackage{graphicx}
\graphicspath{{figures/}}
\usepackage{booktabs}
\usepackage{multirow}
\usepackage{makecell}
\usepackage{pifont}
\usepackage{xcolor}
\usepackage{xspace}
\usepackage{hyperref}
\usepackage{url}
\usepackage{enumitem}
\usepackage{float}
\usepackage{caption}
\usepackage{tabularx}
\newlength{\analysispanelheight}

\newcommand{\benchmark}{\texorpdfstring{M\textsuperscript{2}Weather}{M2Weather}\xspace}

\newcommand{\myparagraph}[1]{\par\noindent\textbf{#1}\hspace{0.5em}\ignorespaces}

\iclrfinalcopy

\title{\benchmark: A Benchmark for Joint Multi-Station and Multi-Variable Weather Forecasting}

\author{
{\bfseries Rongwen Li$^{1}$\thanks{These authors contributed equally.}, Xiao Wang$^{1}$\footnotemark[1], Mingyang Wang$^{1}$\footnotemark[1], Hongwu Liu$^{2}$,}\\
{\bfseries Changjian Chen$^{1}$\thanks{Corresponding to changjianchen@hnu.edu.cn.}, Zhuo Tang$^{1}$, Kenli Li$^{1}$}\\[0.4em]
{\normalfont $^{1}$Hunan University}\\
{\normalfont $^{2}$China Meteorological Administration}
}

\begin{document}
\maketitle
\pagestyle{plain}
\thispagestyle{plain}

\begin{abstract}
Station weather forecasting is fundamentally shaped by both complex spatial dependencies across stations and strong physical coupling among weather variables. However, existing studies often consider these relationships separately and use different datasets and experimental settings, hindering systematic assessment of their individual and joint contributions. 
In this paper, we introduce \benchmark, a benchmark for joint multi-station and multi-variable weather forecasting.
Through multi-criteria quality control and station stratification, we collect 2,809 high-quality stations with 5 physically coupled weather variables across three spatial scales: France, Europe, and Global. This multi-scale design lets us examine whether conclusions persist from national to global station networks. 
We also introduce unified training and evaluation protocols to enable fair comparison of different station--variable modeling paradigms.
To further examine the benefits of modeling station-variable relationships, we design a lightweight, plug-and-play adapter.
With a trained weather forecasting model, this adapter can introduce missing station or variable relationships without retraining the model. This enables fair and efficient investigation of station--variable relationships.
Systematic evaluation of 16 representative models shows the benefits of jointly modeling station and variable relationships. 
Completing missing relationships further reduces MSE for all adapted models on all three datasets. 
Together, these results identify the complementary information across stations and variables as an important resource for improving station weather forecasting. Our code can be obtained at https://github.com/hnu-vis/M2-Weather. 
\end{abstract}

\section{Introduction}
\label{sec:introduction}

Station weather forecasting is a fundamental task in meteorology~\citep{zhu2023weather2k}, with important applications in agriculture, energy scheduling, transportation, and disaster early warning~\citep{allen2025aardvark}.
However, accurate forecasting remains challenging because weather dynamics exhibit both complex spatial dependencies and multivariate coupling.
On the one hand, geographical location, the propagation of atmospheric processes, and regional climate jointly give rise to substantial interactions across weather stations \citep{wu2023corrformer}. 
On the other hand, weather variables, such as temperature, pressure, humidity, and wind speed, do not evolve independently; they continuously interact through atmospheric dynamics and thermodynamics \citep{ma2023histgnn}. Therefore, station weather forecasting is naturally a prediction problem involving interactions among both \emph{multiple stations} and \emph{multiple variables}, where observations carry information about conditions at neighboring stations as well as related weather variables. Their usefulness may also vary with the spatial extent of the station network, motivating evaluation at multiple scales.

\begin{figure}
 \centering
 \includegraphics[width=\linewidth]{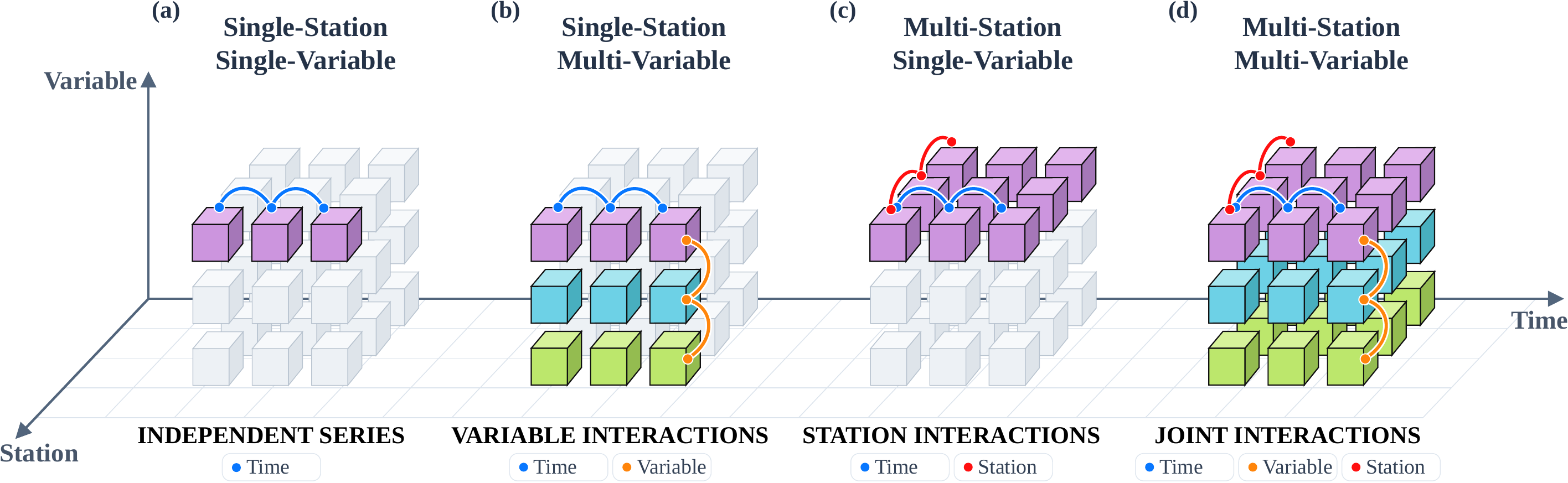}
 \caption{Four station--variable forecasting paradigms: (a) SSSV models each series independently; (b) SSMV captures interactions among variables within a station; (c) MSSV captures interactions across stations for one variable; and (d) MSMV models both. All four use temporal history, but differ in the station and variable context available to each forecast.}
 \label{fig:intro}
\end{figure}

Despite the close interactions between stations and variables, existing research often treats these two dimensions separately. 
These methods can be classified into three categories: Single-Station Single-Variable (SSSV), Single-Station Multi-Variable (SSMV), and Multi-Station Single-Variable (MSSV) methods.
SSSV methods (Fig.~\ref{fig:intro}(a)) model individual stations and variables independently, as in channel-independent forecasting \citep{nie2023patchtst,zeng2023dlinear}.
SSMV methods (Fig.~\ref{fig:intro}(b)) capture dependencies between variables but typically lack explicit representations of station-level spatial structure \citep{liu2024itransformer,woo2024moirai}.
MSSV methods (Fig.~\ref{fig:intro}(c)) capture spatial relationships across stations while predicting different weather variables separately \citep{wu2023corrformer,xu2025cdpnet}.
Methods jointly modeling Multi-Station Multi-Variable (MSMV, Fig.~\ref{fig:intro}(d)) interactions remain relatively limited \citep{ma2023histgnn}. 
More importantly, these methods are commonly evaluated using different datasets, interaction scopes, target variables, and experimental settings, hindering fair comparisons between modeling paradigms. 
We consequently lack a systematic evaluation for a fundamental question: 
\textbf{How do the individual and joint interactions among stations and variables affect station weather forecasting?}

Systematically investigating this question also places greater demands on data. First, reliable evaluation requires high-quality, complete station observations. 
High missing rates, poor data quality, or unstable station records can obscure underlying MSMV interactions, making model differences reflect data noise rather than forecasting capability. 
Second, weather variables have well-defined physical connections: their interactions are governed jointly by weather dynamics and thermodynamics.
A benchmark for MSMV forecasting should therefore select weather quantities with clear physical coupling, so that variable interactions correspond to actual weather dynamics rather than statistical associations alone. 
Existing station datasets provide valuable foundations for forecasting research \citep{zhu2023weather2k,han2024weather5k}, but data and evaluation settings still lack a unified, systematic design that accounts for these two considerations.

To this end, we introduce \textbf{\benchmark}, a \textbf{Multi-Station and Multi-Variable Benchmark} for station weather forecasting. 
This benchmark contains 2,809 high-quality stations with 5 physically coupled weather variables across three spatial scales: France, Europe, and Global.
We further establish a unified, fair protocol for multi-station, multi-variable forecasting. To examine the effects of station and variable interactions, we develop a lightweight, plug-and-play adapter, which can introduce missing station or variable interactions to a trained forecasting model without retraining.
Experiments of 16 baselines on \benchmark show that MSMV methods achieve the best performance.
With the adapter, we find that adding more missing station--variable interactions consistently improve performance for all adapted baselines.
By adding all station--variable interactions, some non-MSMV methods can even achieve state-of-the-art performance.
These findings support the value of MSMV modeling for station weather forecasting. Overall, our main contributions are summarized as follows:\par\vspace{-12pt}
 

\begin{itemize}[leftmargin=*,topsep=-1pt,itemsep=3pt,parsep=0pt]
    \item \textbf{A Benchmark for Joint Station--Variable Forecasting.}
    We construct a collection of high-quality station datasets with physically coupled weather variables and establish consistent training and evaluation protocols for investigating station--variable relationships in station weather forecasting.

    \item \textbf{Backpropagation-Free, Plug-and-Play Interaction Adapters.}
    We design lightweight station- and variable-interaction components that complement relationships missing from a baseline without retraining, enabling controlled analysis of each interaction dimension.

    \item \textbf{Systematic Evidence for MSMV Modeling.}
    Across the three datasets, the evaluated MSMV baseline achieves the best predictive performance, while every adapted backbone consistently improves with interaction completion. These findings highlight the value of jointly modeling station and variable relationships.
\end{itemize}

\section{Taxonomy of Station Weather Forecasting}
\label{sec:taxonomy}
We classify methods by whether they explicitly model cross-station and cross-variable interactions. Shared parameters or pre-training alone do not imply either interaction.

\subsection{Single-Station Single-Variable Forecasting}
Channel-independent methods learn temporal patterns from individual sequences, often sharing a predictor across channels. DLinear combines seasonal--trend decomposition with linear mappings \citep{zeng2023dlinear}, while xPatch uses exponential decomposition and separate MLP/CNN streams to capture trend and seasonal components \citep{stitsyuk2025xpatch}. PatchTST represents local temporal segments as patches, reducing attention costs and enabling longer look-back windows \citep{nie2023patchtst}. This paradigm also extends to foundation models: Timer unifies heterogeneous data as single-series sequences for autoregressive next-token prediction \citep{liu2024timer}, and TimeMoE scales model capacity through sparse experts and multi-resolution forecasting heads \citep{shi2025timemoe}. These methods provide temporal modeling baselines without explicitly coupling station or variable observations.

\subsection{Single-Station Multi-Variable Forecasting}
Multivariate methods exploit dependencies among observed quantities within a station. iTransformer embeds each variable's history as a token and applies attention across variables \citep{liu2024itransformer}. TQNet combines periodically shifted learnable queries with sample-dependent keys and values to capture variable correlations \citep{lin2025tqnet}. DUET jointly clusters temporal patterns and channels, using frequency-domain relationships and sparse attention to select relevant variables \citep{qiu2025duet}. Among general forecasting models, Moirai supports varying numbers of variables through any-variate attention \citep{woo2024moirai}, while Timer-XL extends next-token prediction to multivariate sequences with TimeAttention that combines cross-variable dependencies and temporal causality \citep{liu2025timerxl}. Applied within individual stations, these methods isolate the contribution of variable interactions without explicitly representing station-level spatial structure.

\subsection{Multi-Station Single-Variable Forecasting}
This setting exploits relationships among irregularly distributed stations for a given weather variable. Corrformer combines temporal autocorrelation and spatial cross-correlation within a multiscale tree to capture weather propagation \citep{wu2023corrformer}. S$^2$Transformer uses intra-subgraph and inter-subgraph attention to capture local and regional relationships while supporting large station networks \citep{chen2026s2transformer}. CDPNet reconstructs a continuous spatial field from discrete observations and models its evolution through calibrated initialization and diffusive difference estimation \citep{xu2025cdpnet}. Complementary approaches reduce computational costs: EasyST distills spatio-temporal graph models into an MLP with spatio-temporal prompts \citep{tang2024easyst}, while STELLA incorporates geographical coordinates and temporal embeddings into a lightweight MLP \citep{fu2025stella}. Together, these methods span explicit spatial interaction and lightweight incorporation of spatio-temporal knowledge.

\subsection{Multi-Station Multi-Variable Forecasting}
Joint modeling captures both within-station coupling and cross-station dependencies. HiSTGNN builds adaptive variable and station graphs with bidirectional exchange between levels~\citep{ma2023histgnn}. DIN combines cross-regional dynamics with higher-order variable interactions~\citep{xu2025din}, while StationPDE evolves a station-derived surface field using wind transport and coupled meteorological dynamics~\citep{wang2026stationpde}. These approaches illustrate distinct routes to explicit MSMV modeling, which remains comparatively underexplored.

\begin{figure}
    \centering
    \includegraphics[width=\linewidth]{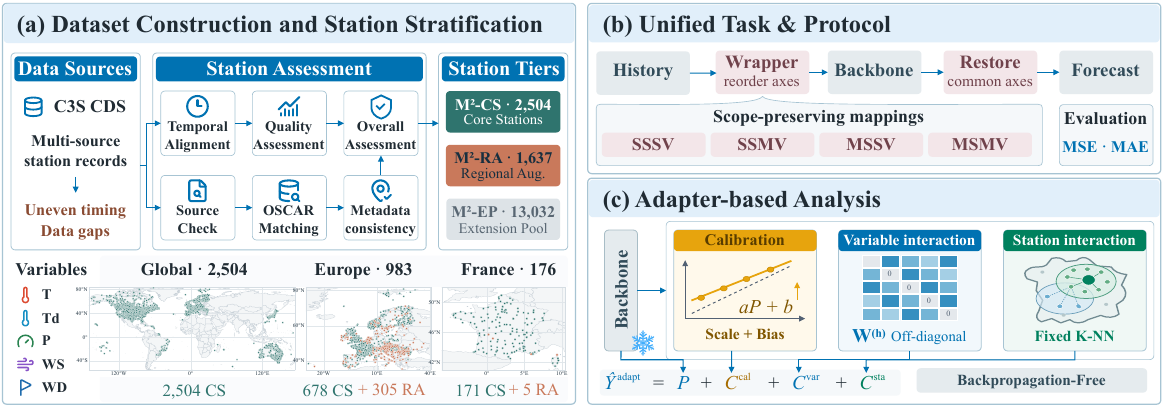}
    \caption{Overview of the \benchmark benchmark and interaction analysis framework. (a) Dataset construction and station stratification based on temporal alignment, quality, provenance, and metadata. (b) Unified protocol for heterogeneous backbones, with model-specific wrappers preserving native interaction structures. (c) Adapter-based analysis on frozen benchmark backbones, with calibration and variable/station interaction corrections fitted by gradient-free ridge regression.}
    \label{fig:pipeline}
\end{figure}

\section{The \benchmark Benchmark}
\label{sec:dataset}

\subsection{Problem Definition}
\label{sec:problem_definition}
For a station network at a given spatial scale, let $N$ denote the number of stations and $V$ the number of weather variables. Let $x_{t,n,v}$ be the observation of variable $v$ at station $n$ and time $t$, and let $L$ and $H$ denote the history length and forecast horizon. The forecasting task is
\begin{equation}
\mathbf X_t=(x_{t-L+1:t,n,v})_{n,v},\quad\mathbf Y_t=(x_{t+1:t+H,n,v})_{n,v},\quad\widehat{\mathbf Y}_t=f_\theta(\mathbf X_t).
\label{eq:forecast_task}
\end{equation}
Here, $\mathbf X_t\in\mathbb R^{L\times N\times V}$ and $\mathbf Y_t\in\mathbb R^{H\times N\times V}$ are the historical input and future target tensors, respectively, while $f_\theta$ maps the former to the forecast $\widehat{\mathbf Y}_t$.

\subsection{Dataset}

\subsubsection{Data Collection}
\label{sec:data_collection}
\benchmark is constructed from global land-surface observations provided by the Copernicus Climate Change Service through the Climate Data Store (CDS) \citep{c3s2026land}. 
The source product integrates station records from National Meteorological Services, aviation observing networks, and historical archives, together with quality flags, report types, and source identifiers.
Despite its broad coverage, stations vary considerably in reporting frequency, variable availability, and temporal continuity, making the raw observations unsuitable for direct multi-station and multi-variable forecasting.
We extract air temperature, dew-point temperature, station pressure, wind speed, and wind direction from 2017 to 2021. 
These variables describe the thermodynamic, moisture, pressure, and dynamical states of the near-surface atmosphere and exhibit meaningful physical coupling \citep{xu2024weathergft,skamarock2019wrf}. 
The resulting candidate pool contains 17,173 stations and 43,824 hourly timestamps.

\subsubsection{Station Assessment and Stratification}
\label{sec:quality_control}
Reliable evaluation of station and variable interactions requires limiting the influence of missing observations, reporting interruptions, and inconsistent station records.
We therefore assess stations along two complementary paths: temporal alignment followed by observation-quality assessment, and source
checking followed by observing-network matching and metadata-consistency
checks. Together, these provide the overall assessment used to organize
stations into tiers, as illustrated in Fig.~\ref{fig:pipeline}(a).

\paragraph{Temporal alignment.}
We first map irregular station reports to the nearest hourly timestamp in Coordinated Universal Time (UTC), retaining the temporally closest record when multiple reports are available for the same station, variable, and hour.
For the released hourly series, we retain finite, quality-controlled observations within 15 minutes of the target hour, providing a standardized temporal basis for subsequent station assessment and dataset construction.

\myparagraph{Quality assessment.}
Station suitability depends not only on overall data availability but also on how missing observations are distributed across variables and time.
We jointly assess the long-term coverage of all five variables, monthly and seasonal stability, consecutive gap lengths, quality-control pass rates, and hourly-alignment rates.
Detailed definitions and thresholds are provided in Appendix~\ref{app:station_selection}.

\myparagraph{Source and network checks.}
We use source identifiers, report types, and data-policy information in the source product to document the provenance of station records.
We then match stations to the World Meteorological Organization (WMO) Observing Systems Capability Analysis and Review (OSCAR)/Surface catalogue \citep{wmo2026oscar}.
This match allows us to check the consistency of station identities, locations, and elevations, while providing information on operating status and observing-network affiliations.
Together with historical observation quality, these source and network checks inform station selection and help interpret the resulting tiers.
Matching rules and metadata definitions are provided in Appendix~\ref{app:station_metadata}.

\myparagraph{Station tiers and dataset settings.}
The overall assessment organizes the 17,173 candidate stations into
three mutually exclusive tiers.
\textbf{M$^2$Weather Core Stations (M$^2$-CS)} contain 2,504 stations with high long-term completeness, stable monthly coverage, and consistent metadata.
\textbf{M$^2$Weather Regional Augmentation Stations (M$^2$-RA)} contain 1,637 additional stations that satisfy the requirements for regional forecasting and improve local spatial coverage. 
The remaining 13,032 stations form the \textbf{M$^2$Weather Extension Pool (M$^2$-EP)} for future tasks and dataset extensions.
Based on these tiers, we construct three dataset settings using the same variables, temporal range, and preprocessing protocol. 
\textbf{Global} contains all 2,504 core stations; \textbf{Europe} contains 678 core and 305 regional augmentation stations, totaling 983; and \textbf{France} contains 171 core and 5 regional augmentation stations, totaling 176.
These settings provide different station configurations for evaluating station and variable interactions under a unified forecasting protocol.

\setlength{\analysispanelheight}{0.235\textwidth}
\begin{figure}[t]
    \centering

    \begin{minipage}[t]{0.40\textwidth}
        \vspace{0pt}
        \begin{minipage}[c][\analysispanelheight][c]{\linewidth}
            \centering
\renewcommand{\arraystretch}{1.15}
\setlength{\tabcolsep}{2pt}

\resizebox{0.98\linewidth}{!}{%
\begin{tabular}{
    @{}
    l
    @{\hspace{0.3em}}
    l
    @{\hspace{1.2em}}
    l
    @{}
}
    \toprule
    Dataset & Stations & Quality rule \\
    \midrule

    Weather2K
        & 2,130
        & \shortstack[l]{$\leq1\%$ missing;\\
          $\leq1\%$ defaults/variable} \\
    \cmidrule(lr){1-3}

    GlobalTempWind
        & 3,850
        & $\leq20\%$ missing data \\
    \cmidrule(lr){1-3}

    Weather5K
        & 5,672
        & $>90\%$ valid hourly records \\
    \cmidrule(lr){1-3}

    M$^2$Weather
        & 2,504/983/176
        & \shortstack[l]{Multi-criteria station\\
          reliability tiers} \\

    \bottomrule
\end{tabular}
}
        \end{minipage}

        \captionsetup{
            font=scriptsize,
            skip=3pt,
            justification=centering
        }
        \captionof{table}{Reported station-screening rules. M$^2$Weather counts denote Global/Europe/France.}
        \label{tab:dataset_comparison}
    \end{minipage}
    \hfill
    %
    \begin{minipage}[t]{0.285\textwidth}
        \vspace{0pt}
        \begin{minipage}[c][\analysispanelheight][c]{\linewidth}
            \centering
            \includegraphics[
                width=\linewidth,
                height=\analysispanelheight,
                keepaspectratio
            ]{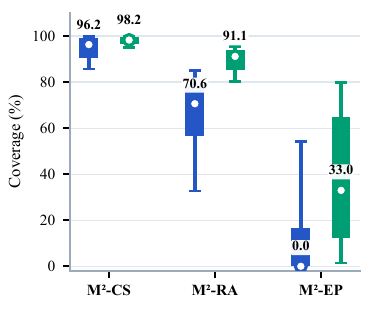}
        \end{minipage}

        \captionsetup{
            font=scriptsize,
            skip=3pt,
            justification=centering
        }
        \captionof{figure}{Pre-imputation coverage:
        minimum in blue and mean in green.}
        \label{fig:coverage_evidence}
    \end{minipage}
    \hfill
    %
    \begin{minipage}[t]{0.285\textwidth}
        \vspace{0pt}
        \begin{minipage}[c][\analysispanelheight][c]{\linewidth}
            \centering
            \includegraphics[
                width=\linewidth,
                height=\analysispanelheight,
                keepaspectratio
            ]{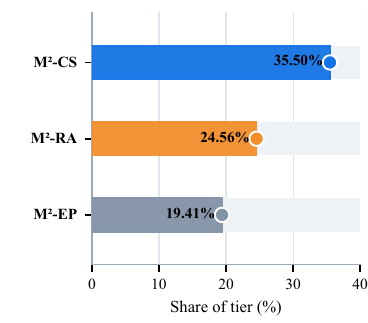}
        \end{minipage}

        \captionsetup{
            font=scriptsize,
            skip=3pt,
            justification=centering
        }
        \captionof{figure}{Verified current GBON affiliation.}
        \label{fig:gbon_evidence}
    \end{minipage}

\end{figure}

\subsubsection{Data Analysis}
\label{sec:data_analysis}

\paragraph{Comparison with existing station datasets.}
Table~\ref{tab:dataset_comparison} compares M$^2$Weather with representative
hourly station-weather datasets
\citep{zhu2023weather2k,wu2023corrformer,han2024weather5k}.
The reported screening rules for these datasets primarily rely on one or two
sequence-based availability thresholds. M$^2$Weather additionally combines
multivariate and temporal quality checks with verified station identities and
observing-network affiliations to define reliability tiers.

\paragraph{Quality and network evidence.}
Fig.~\ref{fig:coverage_evidence} summarizes station coverage before
imputation. The median minimum-variable and mean-variable coverage values are
$96.24\%$ and $98.23\%$ for M$^2$-CS, compared with $70.58\%$ and
$91.11\%$ for M$^2$-RA.
Fig.~\ref{fig:gbon_evidence} provides complementary observing-network
evidence. The proportion of stations with both a verified identity in the
WMO OSCAR/Surface catalogue and current Global Basic Observing Network
(GBON) affiliation decreases from $35.50\%$ in M$^2$-CS to $24.56\%$ in
M$^2$-RA and $19.41\%$ in M$^2$-EP.
Together, these results show that the proposed tiers capture both historical
multivariate usability and externally documented observing-network roles.

\subsection{Unified Training and Evaluation Protocols}
\label{sec:protocols}

\myparagraph{Data splits and standardization.}
For each spatial scale, we split the data chronologically into training, validation, and test sets with a ratio of 60\%/10\%/30\%. Both historical and forecast windows remain entirely within their assigned split. For each variable, we compute the mean and standard deviation over all stations in the training set of the corresponding spatial scale and use these fixed statistics for validation and testing.

\myparagraph{Model wrappers.}
As illustrated in Fig.~\ref{fig:pipeline}(b), we provide unified wrappers that rearrange the station and variable dimensions to match each backbone's native input format and restore predictions to the common output layout. For SSMV models, stations are folded into the batch dimension, whereas for MSSV models, variables are folded into the batch dimension. Parameters are shared across the resulting samples, allowing each trainable baseline to be trained once per dataset and random seed, without training a separate model for each variable or station. This design supports consistent comparison while preserving each model's intended interaction scope.

\myparagraph{Training and hyperparameters.}
We select four variables from the \benchmark datasets for the forecasting task: air temperature, relative humidity, station pressure, and wind speed. All baselines use 48 hours of observations of these variables to predict their values over the next 72 hours. We run all experiments on NVIDIA A100 GPUs (40 GB), with three runs for supervised models using seeds 2024, 2025, and 2026. All trained or fine-tuned backbones use Adam, MSE as the forecasting loss, a common initial learning rate of $10^{-4}$, and the type1 learning-rate decay schedule. Model-specific auxiliary losses and regularization are retained where applicable. Timer, TimeMoE, and Moirai use zero-shot inference. Batch sizes depend on the backbone and spatial scale and are adjusted to available GPU memory by the execution scheduler. Training lasts at most 30 epochs, and checkpoints are selected using each model's validation objective. Early-stopping patience is 10 for France and 5 for Europe and Global. Model-specific hyperparameters, sampling settings, and pretrained inference configurations are detailed in Appendix~\ref{app:model_hyperparameters}.

\myparagraph{Evaluation metrics.}
We report mean squared error (MSE) and mean absolute error (MAE). Let $\Omega$ contain all evaluated forecast-origin, horizon, station, and variable tuples, and let $e_i=\widehat{Y}_i-Y_i$ denote the prediction error for tuple $i$. The metrics are
\begin{equation}
\operatorname{MSE}=\frac{1}{|\Omega|}\sum_{i\in\Omega}e_i^2,\qquad
\operatorname{MAE}=\frac{1}{|\Omega|}\sum_{i\in\Omega}|e_i|.
\label{eq:benchmark_metrics}
\end{equation}
Overall metrics are computed in standardized space with equal weight across variables; per-variable MSE and MAE in the same standardized space are reported in Appendix~\ref{app:full_results}. All methods use the same test origins, forecast steps, and stations, with predictions from overlapping windows evaluated separately.

\section{Plug-and-Play Adapters for Station-Variable Interaction Analysis}

To investigate how joint station--variable modeling benefits weather forecasting, we design lightweight, backpropagation-free post-processing adapters that complement interactions missing from existing backbones. The adapters introduce variable and station information directly into the forecast outputs, enabling a backbone to access both interaction dimensions without modifying its architecture or retraining its parameters. Their coefficients admit a closed-form solution, making interaction completion efficient across the diverse model families in \benchmark. The adapter-based interaction analysis is summarized in Fig.~\ref{fig:pipeline}(c).

\myparagraph{Output-level residual correction.}
Let $\mathbf P_t=f_\theta(\mathbf X_t)\in\mathbb R^{H\times N\times V}$ denote the frozen backbone forecast at origin $t$. The adapter operates in the dataset-standardized space and adds calibration, variable-interaction, and station-interaction corrections:
\begin{equation}
\widehat{\mathbf Y}^{\mathrm{adapt}}_t
=\mathbf P_t+\mathbf C^{\mathrm{cal}}_t+\mathbf C^{\mathrm{var}}_t+\mathbf C^{\mathrm{sta}}_t.
\label{eq:adapter_residual}
\end{equation}
The backbone remains fixed in evaluation mode, and the enabled interaction branches process the same original predictions in parallel. We use $C^{\mathrm{cal}}_{t,h,n,v}=a_{h,v}P_{t,h,n,v}+b_{h,v}$, where the learnable scale and bias corrections are shared across stations. This term accounts for each variable's own amplitude and bias errors, allowing the interaction branches to focus on cross-variable and cross-station information rather than having to absorb these basic calibration effects.

\myparagraph{Variable-interaction adapter.}
For each station $n$ and forecast step $h$, we normalize the prediction vector across its $V$ variables and denote the result by $\widetilde{\mathbf P}_{t,h,n,:}$. We then aggregate the normalized predictions of other variables through a learnable off-diagonal mapping:
\begin{equation}
C^{\mathrm{var}}_{t,h,n,v}
=\sum_{u\ne v}W^{(h)}_{u,v}\widetilde P_{t,h,n,u},
\qquad W^{(h)}_{v,v}=0.
\label{eq:variable_adapter}
\end{equation}
Here, $W^{(h)}_{u,v}$ controls how variable $u$ contributes to the correction of variable $v$ at forecast step $h$. The mapping is shared across stations and introduces variable interactions within each station, without exchanging information between stations.

\myparagraph{Station-interaction adapter.}
We construct a fixed set $\mathcal N(n)$ of eight nearest neighbors for each station, excluding itself, using great-circle angular distances $d_{nj}$ computed from geographical coordinates. Let $\tau_n$ be the median neighbor distance with a positive lower bound. The row-normalized adjacency matrix is
\begin{equation}
A_{nj}=\frac{\exp(-d_{nj}/\tau_n)}{\sum_{k\in\mathcal N(n)}\exp(-d_{nk}/\tau_n)},
\qquad j\in\mathcal N(n),
\label{eq:station_adjacency}
\end{equation}
with all other entries zero. For each forecast step and variable, we normalize predictions across stations and denote the resulting vector by $\mathbf Z_{t,h,:,v}$. The station correction combines one- and two-hop spatial contrasts:
\begin{equation}
\mathbf C^{\mathrm{sta}}_{t,h,:,v}
=\sum_{r=1}^{2}\gamma^{(r)}_{h,v}
\bigl(\mathbf A^r\mathbf Z_{t,h,:,v}-\mathbf Z_{t,h,:,v}\bigr).
\label{eq:station_adapter}
\end{equation}
Each $\gamma^{(r)}_{h,v}$ is a learnable coefficient shared across stations. Its magnitude and sign determine the strength and direction of the correction associated with the $r$-hop neighborhood, adapting the influence of local and broader spatial patterns to each variable and forecast lead time. The contrast measures how a station's prediction differs from its surrounding field. This branch exchanges station information separately for each variable, and neither interaction branch mixes forecast steps.

\myparagraph{Closed-form solution.}
With backbone predictions fixed, the adapter correction is linear in all learnable coefficients. Under squared error, fitting therefore reduces to a quadratic problem that can be solved directly by ridge regression. For each forecast-step--variable pair $(h,v)$, we stack $M=|\mathcal T_{\mathrm{train}}|N$ training origin--station samples into a feature matrix $\boldsymbol\Phi_{h,v}$ and construct a residual vector $\mathbf r_{h,v}$ with entries $Y_{t,h,n,v}-P_{t,h,n,v}$, where $\mathcal T_{\mathrm{train}}$ denotes the training forecast origins. Let $\boldsymbol\beta_{h,v}$ collect $a_{h,v}$, $b_{h,v}$, and the enabled interaction coefficients. We solve
\begin{equation}
\boldsymbol\beta_{h,v}^{\star}
=\arg\min_{\boldsymbol\beta}\;
\frac{1}{M}\|\boldsymbol\Phi_{h,v}\boldsymbol\beta-\mathbf r_{h,v}\|_2^2
+\lambda\boldsymbol\beta^\top\mathbf D\boldsymbol\beta,
\label{eq:adapter_fitting}
\end{equation}
whose normal equations yield the following expression when the coefficient matrix is nonsingular:
\begin{equation}
\boldsymbol\beta_{h,v}^{\star}
=\bigl(\boldsymbol\Phi_{h,v}^{\top}\boldsymbol\Phi_{h,v}+M\lambda\mathbf D\bigr)^{-1}
\boldsymbol\Phi_{h,v}^{\top}\mathbf r_{h,v}.
\label{eq:adapter_closed_form}
\end{equation}
Here, $\mathbf D$ leaves the calibration intercept unpenalized, and $\lambda\geq 0$ is selected by overall standardized validation MSE. We solve the normal equations numerically without explicitly forming the displayed inverse; the expression does not assume that every candidate produces an invertible matrix. We accumulate feature sufficient statistics batch by batch and perform a single small regression solve per $(h,v)$ for each candidate $\lambda$, reusing the same statistics across candidates. This yields a lightweight fitting procedure without iterative gradient descent or backpropagation. Appendix~\ref{app:adapter_fitting} details the feature construction, streaming solution, and parameter-efficiency analysis. In practice, branches are fitted according to the interactions to be completed. The full-completion configurations add station interactions to SSMV, variable interactions to MSSV, and both to SSSV, supplying both information dimensions at the output level.

\section{Observations and Findings}
\label{sec:experiments}
We evaluate 16 representative backbones spanning the SSSV, SSMV, MSSV, and MSMV paradigms on France, Europe, and Global (Table~\ref{tab:benchmark_baselines}). They cover linear models, Transformers, graph neural networks, and pretrained time-series foundation models, enabling comparisons across interaction scopes and model families. For MSMV, we evaluate HiSTGNN using its publicly available implementation. The implementations of DIN and StationPDE are not publicly released, so they cannot be included in the reproducible comparison under our unified protocol. Using 48 hours of history, each model predicts the next 72 hours. The main text reports mean standardized MSE, averaged equally over the four target variables; complete per-variable MSE, MAE, and standard deviations are provided in Appendix~\ref{app:full_results}.

\begin{table}[t]
\centering
\caption{Baseline coverage under the four station--variable modeling paradigms in \benchmark.}
\label{tab:benchmark_baselines}
\small
\renewcommand{\arraystretch}{1.15}
\begin{tabular}{@{}p{0.16\linewidth}p{0.80\linewidth}@{}}
\toprule
Paradigm & Baselines \\
\midrule
SSSV & DLinear~\citep{zeng2023dlinear}, xPatch~\citep{stitsyuk2025xpatch}, PatchTST~\citep{nie2023patchtst}, Timer~\citep{liu2024timer}, TimeMoE~\citep{shi2025timemoe} \\
SSMV & iTransformer~\citep{liu2024itransformer}, TQNet~\citep{lin2025tqnet}, DUET~\citep{qiu2025duet}, Moirai~\citep{woo2024moirai}, Timer-XL~\citep{liu2025timerxl} \\
MSSV & Corrformer~\citep{wu2023corrformer}, S$^2$Transformer~\citep{chen2026s2transformer}, CDPNet~\citep{xu2025cdpnet}, EasyST~\citep{tang2024easyst}, STELLA~\citep{fu2025stella} \\
MSMV & HiSTGNN~\citep{ma2023histgnn} \\
\bottomrule
\end{tabular}
\end{table}

\begin{table}[t]
\centering
\caption{MSE ($48\!\rightarrow\!72$ hours). $\Delta$ is the relative change after adaptation; negative values indicate lower error. HiSTGNN is reported without an adapter. Underlining marks the best native backbone; bold marks the best displayed result (ties at four decimals). Per-variable MSE, MAE, and standard deviations are in Appendix~\ref{app:full_results}.}
\label{tab:main_results}
\begingroup
\setlength{\tabcolsep}{1.7pt}
\renewcommand{\arraystretch}{1.0}
\fontsize{8}{8.5}\selectfont
\begin{tabular*}{\linewidth}{@{\extracolsep{\fill}}llrrrrrrrrr@{}}
\toprule
 & & \multicolumn{3}{c}{\textbf{France}} & \multicolumn{3}{c}{\textbf{Europe}} & \multicolumn{3}{c}{\textbf{Global}} \\
\cmidrule(lr){3-5}\cmidrule(lr){6-8}\cmidrule(l){9-11}
Paradigm & Model & Base & +Adapter & $\Delta$ (\%) & Base & +Adapter & $\Delta$ (\%) & Base & +Adapter & $\Delta$ (\%) \\
\midrule
SSSV & DLinear & 0.4977 & 0.4605 & $-7.49$ & 0.4481 & 0.4087 & $-8.79$ & 0.4508 & 0.4172 & $-7.45$ \\
 & xPatch & 0.3974 & 0.3642 & $-8.36$ & 0.3509 & 0.3280 & $-6.53$ & 0.3587 & 0.3371 & $-6.04$ \\
 & PatchTST & 0.3905 & 0.3601 & $-7.80$ & 0.3472 & 0.3266 & $-5.92$ & 0.3540 & 0.3341 & $-5.61$ \\
 & Timer & 0.5211 & 0.4445 & $-14.70$ & 0.4392 & 0.3845 & $-12.47$ & 0.4547 & 0.3986 & $-12.33$ \\
 & TimeMoE & 0.4354 & 0.3845 & $-11.69$ & 0.3798 & 0.3435 & $-9.56$ & 0.3943 & 0.3559 & $-9.73$ \\
\midrule
SSMV & iTransformer & 0.3502 & 0.3335 & $-4.79$ & 0.3256 & 0.3120 & $-4.18$ & 0.3233 & 0.3119 & $-3.51$ \\
 & TQNet & 0.3704 & 0.3484 & $-5.93$ & 0.3472 & 0.3274 & $-5.70$ & 0.3506 & 0.3333 & $-4.94$ \\
 & DUET & 0.4052 & 0.3737 & $-7.77$ & 0.3905 & 0.3646 & $-6.63$ & 0.3640 & 0.3420 & $-6.06$ \\
 & Moirai & 0.4717 & 0.3817 & $-19.08$ & 0.4124 & 0.3438 & $-16.63$ & 0.4326 & 0.3585 & $-17.13$ \\
 & Timer-XL & 0.3483 & 0.3331 & $-4.36$ & 0.3132 & 0.3031 & $-3.22$ & 0.3177 & 0.3079 & $-3.10$ \\
\midrule
MSSV & Corrformer & 0.3684 & 0.3417 & $-7.24$ & 0.3099 & 0.2988 & $-3.60$ & 0.3189 & 0.3079 & $-3.44$ \\
 & S$^2$Transformer & 0.3776 & 0.3650 & $-3.33$ & 0.3332 & 0.3252 & $-2.39$ & 0.3158 & 0.3148 & $-0.32$ \\
 & CDPNet & 0.3311 & 0.3248 & $-1.91$ & 0.2940 & \textbf{0.2877} & $-2.15$ & 0.2985 & \textbf{0.2961} & $-0.81$ \\
 & EasyST & 0.3442 & 0.3398 & $-1.27$ & 0.3108 & 0.3086 & $-0.71$ & 0.3101 & 0.3087 & $-0.45$ \\
 & STELLA & 0.3878 & 0.3690 & $-4.84$ & 0.3732 & 0.3520 & $-5.68$ & 0.4285 & 0.3931 & $-8.26$ \\
\midrule
MSMV & HiSTGNN & \textbf{\underline{0.3079}} & -- & -- & \underline{0.2908} & -- & -- & \textbf{\underline{0.2961}} & -- & -- \\
\bottomrule
\end{tabular*}
\endgroup
\end{table}

\subsection{Does Joint Station--Variable Modeling Improve Forecasting?}
\label{sec:overall_results}
\label{sec:paradigm_results}
\begin{figure}[!t]
\centering
\includegraphics[width=\linewidth]{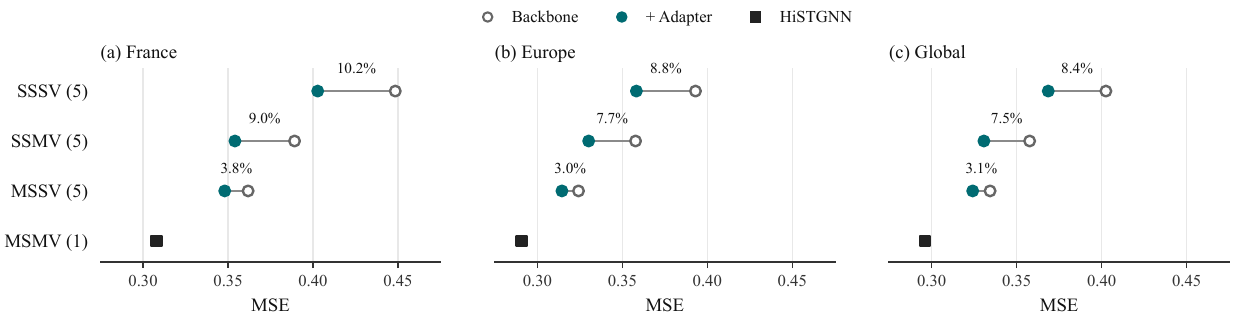}
\caption{MSE by modeling paradigm, averaged over models. Connected markers compare native and adapted predictions; percentages denote reductions in group means. Parentheses indicate group sizes. HiSTGNN is the native MSMV reference, without an adapter.}
\label{fig:paradigm_results}
\end{figure}

We first compare the native backbone results in Table~\ref{tab:main_results} and Figure~\ref{fig:paradigm_results}. Across all three datasets, SSSV has the highest group-mean MSE, whereas the native MSMV reference, HiSTGNN, achieves the lowest MSE among all 16 backbones. On France, for example, the SSSV mean is 0.4484, compared with 0.3892 for SSMV, 0.3618 for MSSV, and 0.3079 for HiSTGNN. The same ordering holds on Europe and Global. This consistent pattern supports the value of jointly exploiting station and variable relationships over treating each series independently.
Comparing the two intermediate paradigms reveals a further pattern: MSSV outperforms SSMV in group-mean MSE at every spatial scale. Their respective means are 0.3618 versus 0.3892 on France, 0.3242 versus 0.3578 on Europe, and 0.3344 versus 0.3576 on Global. These results suggest that information exchanged across stations may be particularly valuable for station weather forecasting. HiSTGNN is the only dedicated MSMV model among the evaluated baselines. Its performance, together with the interaction-completion results below, motivates broader investigation of architectures that explicitly model both station and variable relationships.

\subsection{Does Completing Missing Interactions Improve Forecasting?}
\label{sec:interaction_results}
Table~\ref{tab:main_results} shows that the adapter improves all 15 SSSV, SSMV, and MSSV backbones on every dataset: all 45 model--dataset pairs achieve lower MSE, with relative reductions of 0.32--19.08\%. For the 12 stochastic backbones with three independent runs, seed-level paired tests remain significant after within-dataset Holm correction in 34 of 36 comparisons (Appendix~\ref{app:significance_tests}). Thus, interaction completion benefits diverse backbones while keeping their parameters fixed, providing complementary evidence for jointly modeling the station and variable dimensions.
The group-level results in Figure~\ref{fig:paradigm_results} further reveal where these benefits are largest. SSSV achieves the greatest relative decrease in group-mean MSE at each scale: 10.2\%, 8.8\%, and 8.4\% on France, Europe, and Global. SSMV follows with reductions of 9.0\%, 7.7\%, and 7.5\%, exceeding the corresponding MSSV reductions of 3.8\%, 3.0\%, and 3.1\%. The larger gains for the station-independent groups are consistent with the native comparisons above, suggesting substantial value in adding spatial information. Appendix~\ref{app:model_results} expands the group summaries into individual backbone comparisons.

To examine whether these gains persist without calibration, we disable the calibration branch and refit only the enabled interaction branches on France and Global. Table~\ref{tab:interaction_ablation} in Appendix~\ref{app:interaction_ablation} shows that this interaction-only variant improves all 30 evaluated model--dataset pairs. For example, Moirai's France MSE decreases from 0.4717 to 0.4094 without calibration, and to 0.3817 with the full adapter. Improvements therefore persist when the correction contains only interaction terms, while calibration provides additional benefits. Together, the native comparisons and frozen-backbone results support exploiting both station and variable information for accurate station forecasts.
Prediction examples on Europe (Appendix~\ref{app:prediction_visualization}, Figures~\ref{fig:prediction_t}--\ref{fig:prediction_p}) provide a complementary view: adapted forecasts better track sustained temperature increases, humidity decreases, and rising wind speeds, while most pressure forecasts align more closely with observations. These examples illustrate improvements in forecast levels and trends across backbones.

\section{Conclusion}
\label{sec:conclusion}
We introduced \benchmark to systematically assess the individual and joint roles of station and variable interactions in weather forecasting. The benchmark combines quality-controlled datasets at three spatial scales with unified protocols for comparing four modeling paradigms. Our lightweight, plug-and-play adapters complement missing interactions in frozen backbone forecasts through closed-form ridge regression, requiring no backpropagation. Evaluations of 16 representative models support joint station--variable modeling and highlight the value of spatial information. All 15 adapted backbones achieve lower MSE on all three datasets, and interaction-only ablations show that gains persist without calibration. Together, these findings call for greater attention to forecasting architectures that explicitly capture both station and variable interactions. \benchmark provides a foundation for extending this analysis to richer joint architectures, additional forecast horizons, and extreme-weather conditions.

\section*{AI Use Statement}
Generative AI assisted the authors with preliminary exploration, literature review, and manuscript polishing and editing. The authors remain fully responsible for the accuracy of the text, the cited references, and the reported research.

\bibliographystyle{iclr2027_conference}
\bibliography{main}

\clearpage
\appendix

\section*{Appendix Overview}
This appendix provides supplementary details on dataset construction, benchmark implementation, and experimental results. Section~\ref{app:data_details} covers quality criteria and station tiering, station identity and provenance with network matching, imputation masks and target-variable construction, and spatial subset definitions with additional statistics. Section~\ref{app:experimental_details} presents model hyperparameters across datasets, adapter fitting, and parameter efficiency. Section~\ref{app:additional_results} presents comprehensive forecasting results, ablation and sensitivity analyses, computational efficiency evaluations, and qualitative prediction visualizations.

\section{Dataset Construction Details}
\label{app:data_details}
\subsection{Quality Criteria and Station-Tiering Thresholds}
\label{app:station_selection}

\paragraph{Physical rationale for variable selection.}
We select air temperature ($T$), dew-point temperature ($T_d$),
station pressure ($p$), wind speed ($s$), and wind-from direction
($\phi$) to describe complementary aspects of the near-surface
thermal, moisture, pressure, and flow states. These aspects are
physically connected: pressure gradients contribute to wind
evolution, winds transport heat and moisture between locations,
and changes in water vapour can be accompanied by latent heating.
WRF represents these processes through coupled momentum,
thermodynamic, and moisture equations \citep{skamarock2019wrf},
while WeatherGFT uses their coupled evolution to guide
physics-based forecasting \citep{xu2024weathergft}.
The following simplified atmospheric balances illustrate these
interactions:
\begin{align}
\frac{D\mathbf{u}_h}{Dt}
  + f\hat{\mathbf{k}}\times\mathbf{u}_h
  &= -\frac{1}{\rho}\nabla_h p+\mathbf{F}_h,
  \label{eq:app_momentum}\\
\frac{\partial\rho}{\partial t}
  +\nabla\!\cdot(\rho\mathbf{u})
  &=0,
  \label{eq:app_continuity}\\
c_p\frac{DT}{Dt}
  -\frac{1}{\rho}\frac{Dp}{Dt}
  &=Q_{\mathrm{diab}},
  \label{eq:app_thermodynamics}\\
\frac{Dq}{Dt}
  &=S_q,
  \qquad
  p=\rho R_dT_v,\quad T_v\simeq T(1+0.61q).
  \label{eq:app_moisture}
\end{align}
Here $\mathbf{u}_h=(u,v)$, $\mathbf{u}=(u,v,w)$, and
$D/Dt=\partial_t+\mathbf{u}\cdot\nabla$.
The material derivatives capture wind-driven transport of
temperature and moisture; the pressure-gradient term links
pressure to the flow; and moisture changes associated with
phase transitions also affect the heating term
$Q_{\mathrm{diab}}$.

The selected station observations provide corresponding
near-surface quantities:
\begin{align}
(u,v)&=(-s\sin\phi,\,-s\cos\phi),&
e&=e_s(T_d), \nonumber\\
\mathrm{RH}&=100\frac{e_s(T_d)}{e_s(T)},&
q&=\frac{\varepsilon e}
        {p-(1-\varepsilon)e},
\label{eq:app_observed_variables}
\end{align}
where $e_s$ is saturation vapour pressure, relative humidity is expressed as a percentage, and $\varepsilon=R_d/R_v$. Wind speed and direction jointly
describe the horizontal flow, whereas temperature, dew point,
and pressure jointly characterize near-surface moisture.
Although station observations alone do not resolve the full
three-dimensional state required to integrate the atmospheric
equations, they capture connected aspects of the same physical
system. These local thermodynamic relationships and
wind-mediated links between locations motivate evaluating
variable and station interactions together.

\paragraph{Hourlyization and counting convention.}
Reports are assigned to their nearest UTC hour, with exact half-hour timestamps rounded forward. Reports mapped outside their source calendar month are discarded, and duplicate reports for the same station, variable, and hour are resolved by retaining the report closest to the target hour. Coverage and gap statistics are computed from these de-duplicated hourly cells before imputation. Quality-control (QC) pass rates and hourly-alignment rates are evaluated separately, preventing observation availability, quality flags, and reporting regularity from being conflated.

\paragraph{Quality metrics.}
The evaluation period contains 43,824 hours and 60 calendar months. For each of the five variables, five-year coverage is defined as the number of occupied hourly cells divided by 43,824. Monthly coverage statistics include the lower quantiles of the minimum, mean, and fourth-highest coverage across variables. Temporal continuity is measured using upper quantiles of monthly maximum gap lengths, including gaps at month boundaries. We additionally compute weighted QC pass rates, the proportion of hourly cells represented by reports within 15 minutes of the target hour, and the longest station-wide outage, where an hour is considered available if any target variable is observed.

A month is metadata-stable when its longitude and latitude spans do not exceed $0.01^\circ$ and its valid elevation span does not exceed 10\,m. A station is marked as having an identity conflict when its five-year coordinate or elevation variation exceeds these limits, or when no valid location is available. Each variable is also required to occur in all four meteorological seasons. All monthly quantiles use linear interpolation over the 60 calendar months.

Stations satisfying all M$^2$-CS criteria form the core tier. After excluding these stations, those satisfying all M$^2$-RA criteria form the regional augmentation tier. The remaining stations are retained in the M$^2$-EP extension pool. This procedure yields 2,504 M$^2$-CS, 1,637 M$^2$-RA, and 13,032 M$^2$-EP stations.

\subsection{Station Identity, Provenance, and Network Matching}
\label{app:station_metadata}

\paragraph{Source provenance.}
Station identities are first linked to the C3S station inventory through their primary station identifiers. All 17,173 candidate stations are represented in the inventory. For the observations retained by the strict hourly processing pipeline, each source identifier is resolved against the official C3S source table. We preserve the associated acknowledgement text, report-type composition, and data-policy information as complementary provenance metadata.

\paragraph{OSCAR/Surface matching.}
We cross-reference the station catalogue with the World Meteorological
Organization (WMO) Observing Systems Capability Analysis and Review
(OSCAR)/Surface catalogue \citep{wmo2026oscar}.
Matching considers all available WMO Integrated Global Observing System
(WIGOS) station identifiers and uses three evidence levels.
A match is accepted when: (1) the complete WIGOS identifier matches exactly; (2) the local identifier agrees, the stations are within 10\,km, and country, station name, or elevation provides additional support; or (3) stations belong to the same country and have either an exact name within 5\,km, a name-similarity score of at least 0.90 within 1\,km, or a separation below 0.25\,km with an elevation difference below 100\,m. Coordinate-only matches within 5\,km are retained as possible candidates but are not used to assign official network attributes.

For accepted matches, we record station identity, location, operating status,
and membership in the Global Basic Observing Network (GBON), Regional Basic
Observing Network (RBON), and Global Climate Observing System (GCOS) Surface
Network (GSN).

Network memberships are taken from the OSCAR/Surface snapshot retrieved on 9 September 2026 and therefore provide current external station context. Historical observation quality, C3S source provenance, and WMO observing-network information together constitute the evidence used to interpret the station tiers.

\begin{table}[t]
\centering
\caption{Station-tiering criteria. A dash indicates that the criterion is not used for the corresponding tier.}
\label{tab:station_tiering_thresholds}
\small
\begin{tabular}{lcc}
\toprule
Criterion & M$^2$-CS & M$^2$-RA \\
\midrule
Five-year coverage of every variable       & $\geq 0.82$ & $\geq 0.20$ \\
Five-year mean coverage                    & --          & $\geq 0.65$ \\
Monthly minimum coverage, $P_{10}$         & $\geq 0.60$ & -- \\
Monthly fourth-highest coverage, $P_{20}$  & --          & $\geq 0.50$ \\
Monthly mean coverage, $P_{20}$            & --          & $\geq 0.50$ \\
Monthly maximum five-variable gap, $P_{90}$& $\leq168$ h & -- \\
Monthly maximum existing-variable gap, $P_{80}$ & --     & $\leq72$ h \\
Five-year weighted QC pass rate            & $\geq 0.98$ & $\geq 0.95$ \\
Monthly QC quantile                         & $P_{10}\geq0.95$ & $P_{20}\geq0.95$ \\
Five-year hourly-alignment rate            & $\geq 0.90$ & $\geq 0.75$ \\
Monthly alignment quantile                  & $P_{10}\geq0.75$ & $P_{20}\geq0.75$ \\
Metadata-stable month ratio                & $\geq 0.90$ & $\geq 0.80$ \\
Maximum station-wide outage                & $\leq2160$ h & $\leq2880$ h \\
All variables present in all seasons       & Required    & Required \\
Identity conflict                          & Not allowed & Not allowed \\
\bottomrule
\end{tabular}
\end{table}

\subsection{Imputation Masks and Target-Variable Construction}
\label{app:imputation_targets}

\paragraph{Short-gap interpolation.}
A missing interval is treated as a short internal gap only when it is bounded by valid observations from the same station and variable and contains no more than 12 consecutive hours. Air temperature, dew-point temperature, station pressure, and wind speed are linearly interpolated between the two endpoints. For wind direction, we interpolate along the shortest circular arc. Given endpoint directions $d_l$ and $d_r$, the angular displacement is

\begin{equation}
    \Delta_d =
    \bigl((d_r-d_l+180^\circ) \bmod 360^\circ\bigr)-180^\circ,
\end{equation}

and an intermediate value at relative position $\alpha\in(0,1)$ is computed as

\begin{equation}
    d(\alpha) = (d_l+\alpha\Delta_d) \bmod 360^\circ.
\end{equation}

This formulation correctly interpolates across the $0^\circ/360^\circ$ boundary.

\paragraph{ERA5-based completion.}
After short-gap interpolation, all remaining missing values, including
internal gaps longer than 12 hours and gaps at sequence boundaries, are
matched at the same UTC hour to the nearest $0.25^\circ$ grid cell in ERA5,
the fifth-generation atmospheric reanalysis produced by the European Centre
for Medium-Range Weather Forecasts. Using reanalysis to complete station records follows established dataset-construction practice: WEATHER-5K also applies short-gap interpolation before using ERA5 to fill remaining gaps \citep{han2024weather5k}.

\begin{align}
    T   &= \mathrm{t2m}-273.15, &
    T_d &= \mathrm{d2m}-273.15, \\
    P   &= \mathrm{sp}/100, &
    WS  &= \sqrt{\mathrm{u10}^2+\mathrm{v10}^2}, \\
    WD  &= \left[
    \operatorname{atan2}(-\mathrm{u10},-\mathrm{v10})
    \frac{180^\circ}{\pi}
    \right] \bmod 360^\circ ,
\end{align}

where $T$ and $T_d$ are measured in degrees Celsius, $P$ in hectopascals, $WS$ in meters per second, and $WD$ is the meteorological wind-from direction.

As safeguards, the implementation can fall back sequentially to a nearby hour at the same ERA5 grid cell, a neighboring grid cell at the same hour, an inverse-distance-weighted estimate from neighboring stations, the nearest valid observation from the same station, and finally the cross-station median. None of these fallback stages were activated in the released datasets: all remaining gaps were filled by the nearest ERA5 grid cell at the corresponding hour.

\paragraph{Value-source masks.}
For every released value, we store a source code in
\texttt{value\_source.npy}. The mask has shape
$[N,5,T]$, corresponding to station, meteorological variable, and time. Its dynamic-variable order is air temperature, dew-point temperature, wind speed, station pressure, and wind direction.

\begin{table}[t]
\centering
\caption{Per-value source codes stored in the imputation mask.}
\label{tab:value_source_codes}
\small
\begin{tabular}{clc}
\toprule
Code & Source & Present in release \\
\midrule
0 & Unresolved missing value                         & No  \\
1 & Quality-controlled exact-hour observation        & Yes \\
2 & Quality-controlled observation within 15 minutes & Yes \\
3 & Same-station interpolation of a gap $\leq12$ h   & Yes \\
4 & Nearest ERA5 grid cell at the same hour          & Yes \\
5 & Same ERA5 grid cell at a nearby hour              & No  \\
6 & Neighboring ERA5 grid cell at the same hour       & No  \\
7 & Neighboring-station estimate                      & No  \\
8 & Nearest valid time from the same station          & No  \\
9 & Cross-station median                              & No  \\
\bottomrule
\end{tabular}
\end{table}

\paragraph{Released variables and arrays.}
Each dataset is stored as a \texttt{float32} array with shape $[N,8,T]$. The first three channels are latitude, longitude, and elevation, repeated along the temporal dimension. The remaining channels are air temperature, dew-point temperature, wind speed, station pressure, and wind direction. Wind direction is retained in its original meteorological representation over $[0^\circ,360^\circ)$ rather than being replaced by wind-vector components.

\paragraph{Four-variable forecasting task.}
After short-gap interpolation and completion of any remaining gaps, we derive relative humidity from the processed air and dew-point temperatures. Before standardization, we compute $z=17.625T_d/(243.04+T_d)-17.625T/(243.04+T)$ with both temperatures in degrees Celsius, clip $z$ to $[-50,50]$, and set $\mathrm{RH}=\operatorname{clip}(100\exp(z),0,100)$. Thus, the released five-variable observations retain dew-point temperature, while the forecasting task uses its derived relative humidity together with air temperature, station pressure, and wind speed. Wind speed and direction describe the magnitude and orientation of the same horizontal wind field. We retain wind direction in the released data but select wind speed as the wind-related target in the present four-variable task; joint prediction of both wind quantities is left for future extensions.

\begin{table}[t]
\centering
\caption{Percentage of values completed by either station interpolation or ERA5.}
\label{tab:imputation_rates}
\small
\begin{tabular}{lrrrrrr}
\toprule
Dataset & $T$ & $T_d$ & $WS$ & $P$ & $WD$ & Overall \\
\midrule
Global & 1.392 & 1.953 & 1.578 & 2.343 & 5.576 & 2.568 \\
Europe & 2.608 & 3.216 & 3.630 & 11.242 & 7.499 & 5.639 \\
France & 0.585 & 0.650 & 0.735 & 1.266 & 1.381 & 0.923 \\
\bottomrule
\end{tabular}
\end{table}

All released arrays contain finite values only. The source masks are retained to support observed-only evaluation, imputation-aware training, and sensitivity analyses of the completion strategy.

\paragraph{Dataset standardization.}
For each spatial scale and target variable $v$, we compute $\mu_v=C_v^{-1}\sum_i x_{i,v}$ and $\sigma_v=\sqrt{\max(C_v^{-1}\sum_i x_{i,v}^{2}-\mu_v^2,10^{-12})}$ over all training station--hour values, with equal weight per value and population variance. The maximum ensures $\sigma_v\geq10^{-6}$. Inputs and targets are transformed as $(x_{i,v}-\mu_v)/\sigma_v$, with these statistics frozen for validation and testing. The completed training arrays, including station-interpolated and ERA5-filled values, contribute to these statistics; relative humidity is derived beforehand.

\paragraph{Temporal boundaries.}
All three spatial scales share the same 43,824-hour UTC timeline and the boundaries in Table~\ref{tab:temporal_splits}. The 60\% and 70\% cut points are truncated to integer indices 26,294 and 30,676. Each 48-hour input and its complete 72-hour target remain within one split, without borrowing history from the preceding split. Counts below precede any training or adapter-fitting subsampling. Overlapping forecasts are evaluated as separate origin--lead instances.
\begin{table}[!htbp]
\centering
\caption{Shared temporal splits for France, Europe, and Global. Start times are inclusive and end times exclusive; all times are UTC. Window counts use a one-hour stride.}
\label{tab:temporal_splits}
\small
\setlength{\tabcolsep}{5pt}
\begin{tabular*}{\linewidth}{@{\extracolsep{\fill}}lllr@{}}
\toprule
Split & Start & End & Valid windows \\
\midrule
Training & 2017-01-01 00:00 & 2020-01-01 14:00 & 26,175 \\
Validation & 2020-01-01 14:00 & 2020-07-02 04:00 & 4,263 \\
Test & 2020-07-02 04:00 & 2022-01-01 00:00 & 13,029 \\
\bottomrule
\end{tabular*}
\end{table}

\subsection{Spatial Subset Definitions and Additional Statistics}
\label{app:spatial_subsets}
Global contains all 2,504 M$^2$-CS stations. 
Europe is defined using a frozen World Geodetic System 1984 (WGS84) polygon
bounded by $32^\circ$W--$70^\circ$E and $34^\circ$N--$82^\circ$N, based on
the physical European region and supplemented with country boundaries from
the Geographical Information System of the Commission (GISCO).
Transcontinental countries are included only through their intersection with the European polygon. The resulting subset contains 678 M$^2$-CS and 305 M$^2$-RA stations. France is the subset assigned the GISCO country code \texttt{FR}, containing 171 M$^2$-CS and 5 M$^2$-RA stations. Spatial membership does not modify the quality tier of a station.

The median nearest-neighbor distances are 42.97\,km, 32.62\,km, and 24.80\,km for Global, Europe, and France, respectively. Their corresponding $P_{10}$--$P_{90}$ ranges are 8.22--122.44\,km, 5.79--83.03\,km, and 1.07--54.04\,km, demonstrating progressively denser station configurations across the three settings.

\section{Benchmark Implementation Details}
\label{app:experimental_details}
\subsection{Model Hyperparameters across Datasets}
\label{app:model_hyperparameters}
To ensure a consistent training comparison, all trained or fine-tuned backbones use Adam, MSE as the forecasting loss, an initial learning rate of $10^{-4}$, and the type1 learning-rate decay schedule across France, Europe, and Global. Model-specific auxiliary objectives and regularization are retained where applicable and are described below. Timer, TimeMoE, and Moirai are evaluated in zero-shot mode and therefore have no local optimization settings.

Batch-size triples below follow the order France/Europe/Global and count temporal windows before wrapper reshaping; each window contains all stations. The listed batch sizes are the actual values used in the experiments. Spatial models use 176, 983, and 2,504 stations, respectively, together with the corresponding coordinates. Training lasts at most 30 epochs, with early-stopping patience 10/5/5. Training-window strides are 1/6/6, while validation and testing use stride 1, so training budgets differ across spatial scales. Architecture settings are shared across datasets unless stated otherwise. Weight decay is zero and gradient clipping is disabled unless specified below. Global automatic mixed-precision (AMP) settings are reported per model, with cross-region differences stated where applicable. Reused deterministic predictions are not treated as independent training runs.

\paragraph{DLinear.}
We use a moving-average window of 25 and shared seasonal and trend linear mappings from 48 to 72 steps (\texttt{individual=False}). Batch sizes are $(2048,384,864)$. AMP is disabled on Global.

\paragraph{xPatch.}
We use patch length 16, stride 8, replicated end padding, EMA trend coefficient 0.3, and RevIN. Seasonal and trend streams process each station--variable series independently and restore the instance scale at output. Batch sizes are $(1024,192,128)$; Global AMP is enabled.

\paragraph{PatchTST.}
Patch length and stride are 16 and 8. The encoder has 3 layers, model dimension 128, 16 attention heads, feed-forward dimension 256, and dropout 0.2. Station--variable series are processed independently with instance normalization and output rescaling. Batch sizes are $(128,24,8)$; Global AMP is disabled.

\paragraph{Timer.}
We use \texttt{thuml/timer-base-84m} for zero-shot univariate forecasting, processing at most 4,096 series per internal chunk. Short contexts are padded to a token, and window normalization uses a safe standard deviation for constant inputs. The model generates 72 steps autoregressively. Inference batch sizes are $(128,24,256)$. No local optimizer, training loss, or early stopping is used, and Global AMP is disabled. Global results for the latter two seeds reuse deterministic predictions.

\paragraph{TimeMoE.}
We use \texttt{Maple728/TimeMoE-50M} for zero-shot univariate forecasting, with internal chunks of 1,024 series. Each window is normalized by its own mean and standard deviation; 72 new steps are generated and inverse-transformed. Inference batch sizes are $(4,12,12)$. There is no local training, and Global AMP is disabled. Global results for the latter two seeds reuse deterministic predictions.

\paragraph{iTransformer.}
We use 4 encoder layers, model dimension 256, 8 heads, feed-forward dimension 512, factor 3, and dropout 0.1. Each station's four variables are modeled through inverted variable tokens, with instance normalization and output rescaling. Batch sizes are $(128,24,16)$; Global AMP is disabled.

\paragraph{TQNet.}
We use a 24-hour cycle, the MLP variant, RevIN, dropout 0.5, learnable periodic queries, and 4-head variable attention. The cycle phase is the forecast-start index modulo 24; time-marker column 2 is only a fallback. Batch sizes are $(1024,192,144)$; Global AMP is disabled.

\paragraph{DUET.}
We use model dimension 256, 8 heads, 2 encoder layers, feed-forward dimension 512, hidden size 128, and 4 experts with top-1 routing. Both \texttt{CI} and noisy gating are enabled; dropout and fully connected dropout are 0.2. Each station's four variables are jointly supplied to the model. The forecasting objective is augmented with an expert load-balancing loss of weight 1. Batch sizes are $(256,48,48)$; Global AMP is disabled.

\paragraph{Moirai.}
We use \texttt{Salesforce/moirai-1.0-R-small}, with context length 48 and prediction length 72. At each station, the four target variables are jointly supplied in one inference call, allowing Any-variate Attention to model cross-variable dependencies. Point predictions are obtained by averaging over the sample dimension of the probabilistic forecasts. Inference batch sizes are $(8,16,8)$. There is no local training, and Global AMP is disabled.

\paragraph{Timer-XL.}
We load the pretrained Timer-XL checkpoint and fine-tune all parameters. The effective checkpoint architecture has model dimension 1,024, 8 heads, 8 layers, feed-forward dimension 2,048, and input/output tokens of length 96. Normalization is enabled; covariates and flash attention are disabled. History is left-padded to a 96-step token, and the first 72 steps of the final output token are retained. Gradient norms are clipped at 5. Batch sizes are $(64,12,4)$; Global AMP is disabled. Checkpoint dimensions override generic parser defaults.

\paragraph{Corrformer.}
We use 2 encoder layers, 1 decoder layer, model dimension 256, 8 heads, feed-forward dimension 512, moving-average window 25, temporal and spatial factors of 1, and one decoder TCN layer. Decoder context length is 24. Station hierarchies are $[11,16]$, $[983]$, and $[8,313]$ for France, Europe, and Global. Batch sizes are $(8,1,1)$. Global uses FP32 with saved backward activations offloaded to CPU to reduce GPU memory use.

\paragraph{S$^2$Transformer.}
We use model dimension 256, 8 heads, 2 encoder layers, feed-forward dimension 512, and dropout 0.1. K-means clustering of regional coordinates constructs geographic partitions with 16 and 4 groups, whose memberships vary by dataset. Variables are processed separately. Weight decay is $10^{-4}$, and gradient norms are clipped at 5. Batch sizes are $(384,64,24)$; Global AMP is enabled.

\paragraph{CDPNet.}
We use feature dimension 32, a $32\times32$ grid, and 10 spatial neighbors. Regional coordinates determine the grid extent and station-to-grid interpolation; variables share the spatial network. Weight decay is $10^{-4}$. The first epoch supervises future targets directly; subsequent epochs use a shifted sequence formed from concatenated history and targets, and checkpoint selection starts in the second epoch. Batch sizes are $(256,48,16)$; Global AMP is disabled.

\paragraph{EasyST.}
History, station, time, and transition embeddings each have dimension 64, and the student has 3 MLP layers. A frozen STELLA checkpoint from the same dataset and seed supplies the teacher. Gradient norms are clipped at 5. The forecasting objective additionally includes a teacher-error-gated prediction term weighted by 0.3 (\texttt{teacher\_delta=0.1}) and latent KL regularization weighted by 0.001. Validation excludes the teacher and KL terms. Batch sizes are $(512,96,32)$; Global AMP is enabled.

\paragraph{STELLA.}
We use model dimension 32, 2 layers, dropout 0.2, coordinate-based positional encoding (\texttt{no\_if\_rel}), and residual connections. Variables are folded into the batch dimension and share the spatial network. Weight decay is $5\times10^{-4}$, and gradient norms are clipped at 5. Batch sizes are $(1024,192,512)$; Global AMP is disabled.

\paragraph{HiSTGNN.}
We use four input features, graph-convolution depth 2, convolution channels 32, propagation coefficient 0.05, dropout 0.3, and 3 internal hierarchical layers with station and variable graph convolutions. Weight decay is $10^{-4}$, and gradient norms are clipped at 5. Batch sizes are $(64,20,10)$. AMP is disabled on France and Europe and enabled on Global, where the evaluation batch size is 21.

\subsection{Adapter Fitting and Parameter Efficiency}
\label{app:adapter_fitting}

\paragraph{Feature construction.}
For a fixed forecast step and target variable $(h,v)$, each training origin--station pair $(t,n)$ contributes one row to $\boldsymbol\Phi_{h,v}$. With both interaction branches enabled, the feature and coefficient vectors are
\begin{equation}
\begin{aligned}
\boldsymbol\phi_{t,n}^{(h,v)}
&=\bigl[P_{t,h,n,v},\;1,\;(\widetilde P_{t,h,n,u})_{u\ne v},\;\delta^{(1)}_{t,h,n,v},\;\delta^{(2)}_{t,h,n,v}\bigr]^\top,\\
\boldsymbol\beta_{h,v}
&=\bigl[a_{h,v},\;b_{h,v},\;(W^{(h)}_{u,v})_{u\ne v},\;\gamma^{(1)}_{h,v},\;\gamma^{(2)}_{h,v}\bigr]^\top,
\end{aligned}
\label{eq:adapter_features_appendix}
\end{equation}
where $\delta^{(r)}_{t,h,n,v}=(\mathbf A^r\mathbf Z_{t,h,:,v}-\mathbf Z_{t,h,:,v})_n$. The first feature is the original backbone prediction for calibration; the cross-variable block uses predictions normalized across variables, following Equation~\ref{eq:variable_adapter}. The station features use the station-normalized predictions from Equation~\ref{eq:station_adapter}. All features are fixed before fitting. For an inactive branch, we omit its features and coefficients. Stacking the row vectors $(\boldsymbol\phi_{t,n}^{(h,v)})^\top$ gives $\boldsymbol\Phi_{h,v}\in\mathbb R^{M\times d}$, while the corresponding targets are $r_{t,n}^{(h,v)}=Y_{t,h,n,v}-P_{t,h,n,v}$. Thus, $d=V+3$ for both branches, $d=V+1$ for the variable branch, and $d=4$ for the station branch, each including two calibration coefficients.

\paragraph{Branch normalization and geographic graph.}
Both interaction branches use population variance and the transform $(x-\operatorname{mean}(x))/\sqrt{\operatorname{var}(x)+10^{-5}}$, without learnable affine parameters. The variable branch normalizes over the $V$ predictions at fixed $(t,h,n)$ to obtain $\widetilde{\mathbf P}_{t,h,n,:}$, whereas the station branch normalizes over the $N$ stations at fixed $(t,h,v)$ to obtain $\mathbf Z$. The calibration feature remains the original dataset-standardized prediction. All three branches read the same frozen-backbone output in parallel. Self-variable features are omitted from the regression, enforcing zero diagonal variable-interaction weights.

The fixed eight-neighbor graph uses latitude and longitude only. Distances $d_{nj}$ are haversine central angles in radians. For each station, $\tau_n$ is the fourth-smallest of its eight neighbor distances (the lower median), bounded below by $10^{-8}$ radians. Self-neighbors are excluded, while distinct stations with identical coordinates are retained. Row-wise softmax of $-d_{nj}/\tau_n$ gives $\mathbf A$, which need not be symmetric. The two-hop feature is computed by applying this same fixed graph twice, $\mathbf A(\mathbf A\mathbf Z)-\mathbf Z$.

\paragraph{Analytical solution and batch-wise accumulation.}
Differentiating the quadratic objective in Equation~\ref{eq:adapter_fitting} yields the normal equations and, when the coefficient matrix is nonsingular, the corresponding inverse expression:
\begin{equation}
\begin{aligned}
(\boldsymbol\Phi_{h,v}^{\top}\boldsymbol\Phi_{h,v}+M\lambda\mathbf D)\boldsymbol\beta_{h,v}^{\star}
&=\boldsymbol\Phi_{h,v}^{\top}\mathbf r_{h,v},\\
\boldsymbol\beta_{h,v}^{\star}
&=(\boldsymbol\Phi_{h,v}^{\top}\boldsymbol\Phi_{h,v}+M\lambda\mathbf D)^{-1}
\boldsymbol\Phi_{h,v}^{\top}\mathbf r_{h,v}.
\end{aligned}
\label{eq:adapter_normal_equations}
\end{equation}
With the default calibration branch enabled, $\mathbf D=\operatorname{diag}(1,0,1,\ldots,1)$ leaves only the intercept unpenalized. The candidate set includes $\lambda=0$, so invertibility is not assumed for every candidate. In implementation, we solve the small normal-equation system in CPU float64 using a linear solver, falling back to a least-squares solve if the direct solve raises a linear-algebra error. We never explicitly form the inverse.

The full feature matrix need not be stored. For training batch $b$, let $\boldsymbol\Phi_{h,v}^{(b)}$ and $\mathbf r_{h,v}^{(b)}$ contain its feature rows and residuals. We accumulate only
\begin{equation}
\mathbf G_{h,v}=\sum_b(\boldsymbol\Phi_{h,v}^{(b)})^\top\boldsymbol\Phi_{h,v}^{(b)},
\qquad
\mathbf q_{h,v}=\sum_b(\boldsymbol\Phi_{h,v}^{(b)})^\top\mathbf r_{h,v}^{(b)},
\label{eq:adapter_sufficient_statistics}
\end{equation}
and the total sample count $M$. After all batches have been processed, fitting reduces to $(\mathbf G_{h,v}+M\lambda\mathbf D)\boldsymbol\beta_{h,v}^{\star}=\mathbf q_{h,v}$. These are exactly the full-data normal equations, up to floating-point arithmetic, and require one small solve per $(h,v)$ for a fixed $\lambda$. Candidate regularization strengths reuse the same accumulated statistics and are selected by overall standardized validation MSE; no additional training-data pass is required. Excluding prediction and feature generation, accumulation costs $O(Md^2)$ and the final solve costs $O(d^3)$ per $(h,v)$, with only $O(d^2)$ persistent storage. Since $V=4$, each system has at most seven coefficients. This avoids iterative gradient updates, backpropagation, and optimizer-state storage.

\paragraph{Regularization selection and numerical implementation.}
We select a single $\lambda$ per adapter run from the ordered set $\{0,10^{-7},10^{-6},10^{-5},10^{-4},10^{-3},10^{-2},10^{-1},1\}$. Coefficients are solved separately for each $(h,v)$, but the selected $\lambda$ is shared across all forecast steps, variables, and enabled branches. Selection minimizes aggregate validation MSE over all evaluated origins, stations, steps, and variables; exact ties select the smaller candidate, without a tolerance. The final coefficients use training statistics only, with no training--validation refit.

\paragraph{Parameter efficiency.}
Table~\ref{tab:adapter_parameter_counts} reports parameter counts for representative baselines and the recorded adapter configurations at seed 2024. Counts include independent coefficients that participate in prediction, excluding buffers, duplicate references to shared parameters, and masked adapter coefficients. Adapter counts include calibration but exclude the frozen backbone.
\begin{table}[htbp]
\centering
\caption{Parameter counts across spatial scales. The upper block lists backbone parameters; the lower block lists only additional adapter parameters for the recorded branch configurations.}
\label{tab:adapter_parameter_counts}
\small
\begin{tabular}{llrrr}
\toprule
Paradigm / branch & Model & Global & Europe & France \\
\midrule
\multicolumn{5}{l}{\textit{Backbone parameters}} \\
SSSV & xPatch & 71,076 & 71,076 & 71,076 \\
SSSV & PatchTST & 455,112 & 455,112 & 455,112 \\
SSMV & iTransformer & 2,139,976 & 2,139,976 & 2,139,976 \\
SSMV & TQNet & 596,840 & 596,840 & 596,840 \\
MSSV & S$^2$Transformer & 2,404,936 & 2,404,936 & 2,404,936 \\
MSSV & Corrformer & 144,501,772 & 144,501,772 & 144,501,772 \\
\midrule
\multicolumn{5}{l}{\textit{Additional adapter parameters}} \\
Both & xPatch & 2,016 & 2,016 & 2,016 \\
Both & PatchTST & 2,016 & 2,016 & 2,016 \\
Variable & S$^2$Transformer & 1,440 & 1,440 & 1,440 \\
Variable & Corrformer & 1,440 & 1,440 & 1,440 \\
Station & iTransformer & 1,152 & 1,152 & 1,152 \\
Station & TQNet & 1,152 & 1,152 & 1,152 \\
\bottomrule
\end{tabular}
\end{table}

Calibration contributes $2HV$ coefficients, variable interactions contribute $HV(V-1)$, and the two-hop station branch contributes $2HV$. With $H=72$ and $V=4$, station-only, variable-only, and joint adapters therefore contain 1,152, 1,440, and 2,016 coefficients, respectively. These counts are independent of the number of stations, so they remain unchanged across France ($N=176$), Europe ($N=983$), and Global ($N=2{,}504$). For iTransformer and TQNet, the station-only adapter adds just 1,152 parameters, or 0.054\% and 0.193\% of their respective backbone sizes. The joint adapter adds 2.84\% of xPatch's parameters and 0.44\% of PatchTST's; the variable-only adapter adds 0.060\% of S$^2$Transformer's and 0.0010\% of Corrformer's. This station-shared parameterization supports interaction modeling with minimal additional model size.

\section{Additional Results}
\label{app:additional_results}
\subsection{Reporting Conventions and Adapter Configurations}
\label{app:result_conventions}
All comparisons use 48 hours of observations to predict 72 hours. The following tables report per-variable MSE and MAE. Entries are means $\pm$ sample standard deviations across the recorded seeds 2024, 2025, and 2026 (ddof $=1$). Standard deviations describe variation between run-level metrics, not the distribution of individual forecast errors. Values are rounded to four decimals; $0.0000$ can denote a smaller nonzero deviation. Timer, TimeMoE, and Moirai use deterministic pretrained inference, so repeated values do not represent independently trained backbones.

For both MSE and MAE, $\Delta=100(E_{\mathrm{adapt}}/E_{\mathrm{base}}-1)$ denotes relative change: negative values indicate improvement and positive values indicate deterioration. Percentages are calculated before rounding. Group means weight models equally; group-level changes compare these means rather than averaging model-wise percentages. HiSTGNN is always shown without adaptation and is excluded from adapter improvement counts.

\myparagraph{Evaluated branches.}
Model groups follow Table~\ref{tab:benchmark_baselines}.
DLinear, xPatch, PatchTST, Timer, and TimeMoE use both interaction branches;
iTransformer, TQNet, DUET, Moirai, and Timer-XL use the station branch;
Corrformer, S$^2$Transformer, CDPNet, EasyST, and STELLA use the variable branch.
Full adapters additionally include calibration.

\subsection{Seed-Level Significance Tests}
\label{app:significance_tests}
We assess whether the MSE reductions in Table~\ref{tab:main_results} are consistent across random seeds. For each stochastic backbone and dataset, we apply a two-sided paired $t$-test to the overall standardized MSE from seeds 2024, 2025, and 2026, pairing the native and adapted results from the same seed. The null hypothesis is that the mean paired difference is zero. We then control the family-wise error rate within each dataset using Holm correction over its 12 valid comparisons, with $\alpha=0.05$. Timer, TimeMoE, and Moirai are excluded because their deterministic pretrained predictions are reused across seeds, while HiSTGNN has no adapted counterpart.

\begin{table}[H]
\centering
\caption{Seed-level significance tests for the MSE reductions in Table~\ref{tab:main_results}. Entries are two-sided paired $t$-test $p$-values after Holm correction within each dataset; bold values are significant at $\alpha=0.05$. Each test uses three paired seeds ($df=2$).}
\label{tab:significance_tests}
\begingroup
\small
\setlength{\tabcolsep}{7pt}
\renewcommand{\arraystretch}{1.08}
\begin{tabular*}{\linewidth}{@{\extracolsep{\fill}}llccc@{}}
\toprule
Paradigm & Model & France & Europe & Global \\
\midrule
SSSV & DLinear & $\mathbf{<0.001}$ & $\mathbf{<0.001}$ & $\mathbf{0.0044}$ \\
     & xPatch & $\mathbf{<0.001}$ & $\mathbf{<0.001}$ & $\mathbf{0.0058}$ \\
     & PatchTST & $\mathbf{<0.001}$ & $\mathbf{<0.001}$ & $\mathbf{<0.001}$ \\
\midrule
SSMV & iTransformer & $\mathbf{<0.001}$ & $\mathbf{<0.001}$ & $\mathbf{0.0058}$ \\
     & TQNet & $\mathbf{<0.001}$ & $\mathbf{0.0069}$ & $\mathbf{0.0313}$ \\
     & DUET & $\mathbf{0.0043}$ & $\mathbf{0.0193}$ & $\mathbf{0.0128}$ \\
     & Timer-XL & $\mathbf{<0.001}$ & $\mathbf{0.0069}$ & $\mathbf{<0.001}$ \\
\midrule
MSSV & Corrformer & $\mathbf{0.0043}$ & $\mathbf{0.0045}$ & $\mathbf{0.0035}$ \\
     & S$^2$Transformer & $\mathbf{<0.001}$ & $\mathbf{0.0045}$ & $\mathbf{0.0042}$ \\
     & CDPNet & $0.0808$ & $\mathbf{0.0193}$ & $\mathbf{0.0313}$ \\
     & EasyST & $\mathbf{<0.001}$ & $\mathbf{0.0013}$ & $\mathbf{0.0038}$ \\
     & STELLA & $\mathbf{<0.001}$ & $\mathbf{0.0045}$ & $0.0547$ \\
\bottomrule
\end{tabular*}
\endgroup
\end{table}

After within-dataset correction, 11 of 12 testable backbones are significant on France, all 12 are significant on Europe, and 11 of 12 are significant on Global. The exceptions are CDPNet on France ($p_{\mathrm{Holm}}=0.0808$) and STELLA on Global ($p_{\mathrm{Holm}}=0.0547$). If all 36 tests across the three datasets are treated as a single Holm family, 30 remain significant; Europe DUET and CDPNet and Global TQNet and CDPNet additionally exceed 0.05. 

\subsection{Complete Per-Variable MSE and MAE}
\label{app:full_results}
Tables~\ref{tab:france_t}--\ref{tab:global_p} provide one table for every dataset--variable pair. All 180 adapted model--dataset--variable combinations improve MSE. At the displayed precision of the relative changes, MAE decreases in 56/60, 54/60, and 55/60 combinations on France, Europe, and Global, respectively. One additional Europe comparison (TQNet on station pressure) rounds to $0.00\%$ and is reported as unchanged at this precision. The remaining 4, 5, and 5 comparisons show MAE increases; the adapter is fitted and selected using squared error.
\begin{table}[!htbp]
\centering
\caption{\textbf{France --- Air temperature.} MSE and MAE (mean $\pm$ sample standard deviation). $\Delta$ (\%) is signed relative change; negative is better.}
\label{tab:france_t}
\begingroup
\fontsize{8}{9}\selectfont
\setlength{\tabcolsep}{2pt}
\renewcommand{\arraystretch}{1.07}
\begin{tabular*}{\linewidth}{@{\extracolsep{\fill}}lrrrrrr@{}}
\toprule
 & \multicolumn{3}{c}{\textbf{MSE $\downarrow$}} & \multicolumn{3}{c}{\textbf{MAE $\downarrow$}} \\
\cmidrule(lr){2-4}\cmidrule(l){5-7}
Model & Base & +Adapter & $\Delta$ (\%) & Base & +Adapter & $\Delta$ (\%) \\
\midrule
DLinear & $0.3146\!\pm\!0.0003$ & $0.2928\!\pm\!0.0000$ & $-6.94$ & $0.4411\!\pm\!0.0005$ & $0.4259\!\pm\!0.0000$ & $-3.44$ \\
xPatch & $0.1977\!\pm\!0.0004$ & $0.1921\!\pm\!0.0003$ & $-2.81$ & $0.3371\!\pm\!0.0003$ & $0.3342\!\pm\!0.0003$ & $-0.89$ \\
PatchTST & $0.1987\!\pm\!0.0005$ & $0.1861\!\pm\!0.0005$ & $-6.36$ & $0.3393\!\pm\!0.0005$ & $0.3298\!\pm\!0.0005$ & $-2.79$ \\
Timer & $0.3078\!\pm\!0.0000$ & $0.2718\!\pm\!0.0000$ & $-11.67$ & $0.4235\!\pm\!0.0000$ & $0.4034\!\pm\!0.0000$ & $-4.74$ \\
TimeMoE & $0.2126\!\pm\!0.0000$ & $0.1976\!\pm\!0.0000$ & $-7.07$ & $0.3476\!\pm\!0.0000$ & $0.3377\!\pm\!0.0000$ & $-2.83$ \\
\midrule
iTransformer & $0.1693\!\pm\!0.0007$ & $0.1639\!\pm\!0.0008$ & $-3.18$ & $0.3099\!\pm\!0.0007$ & $0.3061\!\pm\!0.0007$ & $-1.21$ \\
TQNet & $0.1881\!\pm\!0.0006$ & $0.1822\!\pm\!0.0006$ & $-3.12$ & $0.3271\!\pm\!0.0006$ & $0.3239\!\pm\!0.0005$ & $-0.98$ \\
DUET & $0.2082\!\pm\!0.0065$ & $0.2008\!\pm\!0.0040$ & $-3.56$ & $0.3473\!\pm\!0.0053$ & $0.3428\!\pm\!0.0040$ & $-1.31$ \\
Moirai & $0.2326\!\pm\!0.0000$ & $0.1992\!\pm\!0.0000$ & $-14.36$ & $0.3602\!\pm\!0.0000$ & $0.3393\!\pm\!0.0000$ & $-5.80$ \\
Timer-XL & $0.1687\!\pm\!0.0003$ & $0.1620\!\pm\!0.0002$ & $-3.97$ & $0.3087\!\pm\!0.0002$ & $0.3038\!\pm\!0.0002$ & $-1.58$ \\
\midrule
Corrformer & $0.1787\!\pm\!0.0043$ & $0.1658\!\pm\!0.0067$ & $-7.21$ & $0.3202\!\pm\!0.0030$ & $0.3103\!\pm\!0.0053$ & $-3.12$ \\
S$^2$Transformer & $0.1773\!\pm\!0.0011$ & $0.1720\!\pm\!0.0009$ & $-2.95$ & $0.3241\!\pm\!0.0010$ & $0.3194\!\pm\!0.0009$ & $-1.44$ \\
CDPNet & $0.1636\!\pm\!0.0068$ & $0.1577\!\pm\!0.0064$ & $-3.61$ & $0.3101\!\pm\!0.0057$ & $0.3042\!\pm\!0.0057$ & $-1.88$ \\
EasyST & $0.1783\!\pm\!0.0008$ & $0.1743\!\pm\!0.0004$ & $-2.25$ & $0.3221\!\pm\!0.0007$ & $0.3185\!\pm\!0.0003$ & $-1.12$ \\
STELLA & $0.1987\!\pm\!0.0006$ & $0.1915\!\pm\!0.0003$ & $-3.62$ & $0.3421\!\pm\!0.0009$ & $0.3360\!\pm\!0.0006$ & $-1.77$ \\
\midrule
HiSTGNN & $0.1438\!\pm\!0.0063$ & -- & -- & $0.2883\!\pm\!0.0062$ & -- & -- \\
\bottomrule
\end{tabular*}
\endgroup
\end{table}
\begin{table}[!htbp]
\centering
\caption{\textbf{France --- Wind speed.} MSE and MAE (mean $\pm$ sample standard deviation). $\Delta$ (\%) is signed relative change; negative is better.}
\label{tab:france_ws}
\begingroup
\fontsize{8}{9}\selectfont
\setlength{\tabcolsep}{2pt}
\renewcommand{\arraystretch}{1.07}
\begin{tabular*}{\linewidth}{@{\extracolsep{\fill}}lrrrrrr@{}}
\toprule
 & \multicolumn{3}{c}{\textbf{MSE $\downarrow$}} & \multicolumn{3}{c}{\textbf{MAE $\downarrow$}} \\
\cmidrule(lr){2-4}\cmidrule(l){5-7}
Model & Base & +Adapter & $\Delta$ (\%) & Base & +Adapter & $\Delta$ (\%) \\
\midrule
DLinear & $0.8359\!\pm\!0.0003$ & $0.7586\!\pm\!0.0000$ & $-9.25$ & $0.6746\!\pm\!0.0014$ & $0.6506\!\pm\!0.0000$ & $-3.56$ \\
xPatch & $0.7956\!\pm\!0.0006$ & $0.7054\!\pm\!0.0003$ & $-11.34$ & $0.6394\!\pm\!0.0003$ & $0.6171\!\pm\!0.0003$ & $-3.49$ \\
PatchTST & $0.7811\!\pm\!0.0002$ & $0.7029\!\pm\!0.0001$ & $-10.01$ & $0.6374\!\pm\!0.0004$ & $0.6165\!\pm\!0.0001$ & $-3.28$ \\
Timer & $0.9058\!\pm\!0.0000$ & $0.7490\!\pm\!0.0000$ & $-17.31$ & $0.6968\!\pm\!0.0000$ & $0.6434\!\pm\!0.0000$ & $-7.68$ \\
TimeMoE & $0.8696\!\pm\!0.0000$ & $0.7362\!\pm\!0.0000$ & $-15.34$ & $0.6694\!\pm\!0.0000$ & $0.6328\!\pm\!0.0000$ & $-5.47$ \\
\midrule
iTransformer & $0.7184\!\pm\!0.0003$ & $0.6811\!\pm\!0.0007$ & $-5.20$ & $0.6115\!\pm\!0.0006$ & $0.6140\!\pm\!0.0002$ & $+0.42$ \\
TQNet & $0.7514\!\pm\!0.0009$ & $0.6957\!\pm\!0.0005$ & $-7.41$ & $0.6216\!\pm\!0.0004$ & $0.6120\!\pm\!0.0002$ & $-1.55$ \\
DUET & $0.7989\!\pm\!0.0085$ & $0.7180\!\pm\!0.0046$ & $-10.12$ & $0.6386\!\pm\!0.0023$ & $0.6244\!\pm\!0.0021$ & $-2.23$ \\
Moirai & $0.9476\!\pm\!0.0000$ & $0.7383\!\pm\!0.0000$ & $-22.09$ & $0.6899\!\pm\!0.0000$ & $0.6318\!\pm\!0.0000$ & $-8.42$ \\
Timer-XL & $0.7136\!\pm\!0.0005$ & $0.6838\!\pm\!0.0009$ & $-4.17$ & $0.6100\!\pm\!0.0011$ & $0.6022\!\pm\!0.0002$ & $-1.28$ \\
\midrule
Corrformer & $0.7506\!\pm\!0.0100$ & $0.6931\!\pm\!0.0071$ & $-7.67$ & $0.6259\!\pm\!0.0089$ & $0.6042\!\pm\!0.0016$ & $-3.47$ \\
S$^2$Transformer & $0.6860\!\pm\!0.0002$ & $0.6820\!\pm\!0.0002$ & $-0.59$ & $0.5990\!\pm\!0.0002$ & $0.5970\!\pm\!0.0001$ & $-0.33$ \\
CDPNet & $0.6636\!\pm\!0.0109$ & $0.6486\!\pm\!0.0169$ & $-2.26$ & $0.5818\!\pm\!0.0087$ & $0.5855\!\pm\!0.0088$ & $+0.65$ \\
EasyST & $0.6740\!\pm\!0.0009$ & $0.6697\!\pm\!0.0009$ & $-0.64$ & $0.5996\!\pm\!0.0006$ & $0.5986\!\pm\!0.0005$ & $-0.16$ \\
STELLA & $0.7637\!\pm\!0.0010$ & $0.7150\!\pm\!0.0018$ & $-6.38$ & $0.6324\!\pm\!0.0004$ & $0.6187\!\pm\!0.0009$ & $-2.18$ \\
\midrule
HiSTGNN & $0.6273\!\pm\!0.0052$ & -- & -- & $0.5621\!\pm\!0.0041$ & -- & -- \\
\bottomrule
\end{tabular*}
\endgroup
\end{table}
\begin{table}[!htbp]
\centering
\caption{\textbf{France --- Relative humidity.} MSE and MAE (mean $\pm$ sample standard deviation). $\Delta$ (\%) is signed relative change; negative is better.}
\label{tab:france_rh}
\begingroup
\fontsize{8}{9}\selectfont
\setlength{\tabcolsep}{2pt}
\renewcommand{\arraystretch}{1.07}
\begin{tabular*}{\linewidth}{@{\extracolsep{\fill}}lrrrrrr@{}}
\toprule
 & \multicolumn{3}{c}{\textbf{MSE $\downarrow$}} & \multicolumn{3}{c}{\textbf{MAE $\downarrow$}} \\
\cmidrule(lr){2-4}\cmidrule(l){5-7}
Model & Base & +Adapter & $\Delta$ (\%) & Base & +Adapter & $\Delta$ (\%) \\
\midrule
DLinear & $0.7059\!\pm\!0.0007$ & $0.6835\!\pm\!0.0000$ & $-3.16$ & $0.6742\!\pm\!0.0020$ & $0.6734\!\pm\!0.0000$ & $-0.12$ \\
xPatch & $0.4959\!\pm\!0.0007$ & $0.4671\!\pm\!0.0006$ & $-5.82$ & $0.5305\!\pm\!0.0005$ & $0.5316\!\pm\!0.0005$ & $+0.22$ \\
PatchTST & $0.4882\!\pm\!0.0005$ & $0.4657\!\pm\!0.0005$ & $-4.61$ & $0.5346\!\pm\!0.0001$ & $0.5311\!\pm\!0.0003$ & $-0.67$ \\
Timer & $0.7680\!\pm\!0.0000$ & $0.6644\!\pm\!0.0000$ & $-13.50$ & $0.6638\!\pm\!0.0000$ & $0.6551\!\pm\!0.0000$ & $-1.31$ \\
TimeMoE & $0.5592\!\pm\!0.0000$ & $0.5141\!\pm\!0.0000$ & $-8.07$ & $0.5644\!\pm\!0.0000$ & $0.5635\!\pm\!0.0000$ & $-0.16$ \\
\midrule
iTransformer & $0.4309\!\pm\!0.0005$ & $0.4111\!\pm\!0.0005$ & $-4.58$ & $0.4899\!\pm\!0.0004$ & $0.4887\!\pm\!0.0003$ & $-0.24$ \\
TQNet & $0.4541\!\pm\!0.0005$ & $0.4341\!\pm\!0.0004$ & $-4.41$ & $0.5065\!\pm\!0.0003$ & $0.5043\!\pm\!0.0003$ & $-0.42$ \\
DUET & $0.5122\!\pm\!0.0132$ & $0.4826\!\pm\!0.0133$ & $-5.78$ & $0.5417\!\pm\!0.0089$ & $0.5402\!\pm\!0.0094$ & $-0.28$ \\
Moirai & $0.5998\!\pm\!0.0000$ & $0.5006\!\pm\!0.0000$ & $-16.55$ & $0.5707\!\pm\!0.0000$ & $0.5540\!\pm\!0.0000$ & $-2.93$ \\
Timer-XL & $0.4323\!\pm\!0.0004$ & $0.4114\!\pm\!0.0005$ & $-4.84$ & $0.4921\!\pm\!0.0008$ & $0.4907\!\pm\!0.0003$ & $-0.28$ \\
\midrule
Corrformer & $0.4527\!\pm\!0.0006$ & $0.4202\!\pm\!0.0038$ & $-7.20$ & $0.5015\!\pm\!0.0008$ & $0.4958\!\pm\!0.0023$ & $-1.14$ \\
S$^2$Transformer & $0.4386\!\pm\!0.0007$ & $0.4336\!\pm\!0.0008$ & $-1.14$ & $0.5130\!\pm\!0.0005$ & $0.5090\!\pm\!0.0007$ & $-0.77$ \\
CDPNet & $0.4252\!\pm\!0.0080$ & $0.4218\!\pm\!0.0107$ & $-0.80$ & $0.4926\!\pm\!0.0064$ & $0.4910\!\pm\!0.0049$ & $-0.33$ \\
EasyST & $0.4473\!\pm\!0.0016$ & $0.4393\!\pm\!0.0016$ & $-1.80$ & $0.5189\!\pm\!0.0009$ & $0.5117\!\pm\!0.0011$ & $-1.39$ \\
STELLA & $0.4909\!\pm\!0.0015$ & $0.4736\!\pm\!0.0016$ & $-3.52$ & $0.5357\!\pm\!0.0011$ & $0.5370\!\pm\!0.0011$ & $+0.24$ \\
\midrule
HiSTGNN & $0.3892\!\pm\!0.0054$ & -- & -- & $0.4662\!\pm\!0.0035$ & -- & -- \\
\bottomrule
\end{tabular*}
\endgroup
\end{table}
\begin{table}[!htbp]
\centering
\caption{\textbf{France --- Station pressure.} MSE and MAE (mean $\pm$ sample standard deviation). $\Delta$ (\%) is signed relative change; negative is better.}
\label{tab:france_p}
\begingroup
\fontsize{8}{9}\selectfont
\setlength{\tabcolsep}{2pt}
\renewcommand{\arraystretch}{1.07}
\begin{tabular*}{\linewidth}{@{\extracolsep{\fill}}lrrrrrr@{}}
\toprule
 & \multicolumn{3}{c}{\textbf{MSE $\downarrow$}} & \multicolumn{3}{c}{\textbf{MAE $\downarrow$}} \\
\cmidrule(lr){2-4}\cmidrule(l){5-7}
Model & Base & +Adapter & $\Delta$ (\%) & Base & +Adapter & $\Delta$ (\%) \\
\midrule
DLinear & $0.1346\!\pm\!0.0004$ & $0.1069\!\pm\!0.0000$ & $-20.53$ & $0.2715\!\pm\!0.0013$ & $0.2379\!\pm\!0.0000$ & $-12.38$ \\
xPatch & $0.1005\!\pm\!0.0004$ & $0.0922\!\pm\!0.0003$ & $-8.25$ & $0.2211\!\pm\!0.0006$ & $0.2141\!\pm\!0.0006$ & $-3.16$ \\
PatchTST & $0.0941\!\pm\!0.0006$ & $0.0856\!\pm\!0.0005$ & $-8.99$ & $0.2121\!\pm\!0.0009$ & $0.2044\!\pm\!0.0008$ & $-3.64$ \\
Timer & $0.1030\!\pm\!0.0000$ & $0.0929\!\pm\!0.0000$ & $-9.73$ & $0.2222\!\pm\!0.0000$ & $0.2133\!\pm\!0.0000$ & $-4.03$ \\
TimeMoE & $0.1002\!\pm\!0.0000$ & $0.0901\!\pm\!0.0000$ & $-10.06$ & $0.2174\!\pm\!0.0000$ & $0.2084\!\pm\!0.0000$ & $-4.13$ \\
\midrule
iTransformer & $0.0824\!\pm\!0.0003$ & $0.0777\!\pm\!0.0003$ & $-5.69$ & $0.1934\!\pm\!0.0002$ & $0.1892\!\pm\!0.0002$ & $-2.18$ \\
TQNet & $0.0881\!\pm\!0.0003$ & $0.0818\!\pm\!0.0002$ & $-7.19$ & $0.2023\!\pm\!0.0003$ & $0.1965\!\pm\!0.0004$ & $-2.83$ \\
DUET & $0.1014\!\pm\!0.0035$ & $0.0933\!\pm\!0.0031$ & $-7.99$ & $0.2221\!\pm\!0.0041$ & $0.2151\!\pm\!0.0041$ & $-3.14$ \\
Moirai & $0.1067\!\pm\!0.0000$ & $0.0887\!\pm\!0.0000$ & $-16.90$ & $0.2223\!\pm\!0.0000$ & $0.2062\!\pm\!0.0000$ & $-7.23$ \\
Timer-XL & $0.0786\!\pm\!0.0003$ & $0.0753\!\pm\!0.0003$ & $-4.26$ & $0.1880\!\pm\!0.0005$ & $0.1849\!\pm\!0.0004$ & $-1.63$ \\
\midrule
Corrformer & $0.0916\!\pm\!0.0033$ & $0.0879\!\pm\!0.0027$ & $-3.98$ & $0.2042\!\pm\!0.0039$ & $0.2007\!\pm\!0.0029$ & $-1.69$ \\
S$^2$Transformer & $0.2087\!\pm\!0.0043$ & $0.1725\!\pm\!0.0030$ & $-17.31$ & $0.2338\!\pm\!0.0010$ & $0.2274\!\pm\!0.0014$ & $-2.75$ \\
CDPNet & $0.0721\!\pm\!0.0009$ & $0.0711\!\pm\!0.0005$ & $-1.39$ & $0.1851\!\pm\!0.0040$ & $0.1816\!\pm\!0.0010$ & $-1.88$ \\
EasyST & $0.0773\!\pm\!0.0003$ & $0.0761\!\pm\!0.0003$ & $-1.49$ & $0.1906\!\pm\!0.0003$ & $0.1885\!\pm\!0.0005$ & $-1.09$ \\
STELLA & $0.0978\!\pm\!0.0007$ & $0.0959\!\pm\!0.0005$ & $-1.92$ & $0.2238\!\pm\!0.0018$ & $0.2211\!\pm\!0.0016$ & $-1.22$ \\
\midrule
HiSTGNN & $0.0712\!\pm\!0.0021$ & -- & -- & $0.1812\!\pm\!0.0021$ & -- & -- \\
\bottomrule
\end{tabular*}
\endgroup
\end{table}
\begin{table}[!htbp]
\centering
\caption{\textbf{Europe --- Air temperature.} MSE and MAE (mean $\pm$ sample standard deviation). $\Delta$ (\%) is signed relative change; negative is better.}
\label{tab:europe_t}
\begingroup
\fontsize{8}{9}\selectfont
\setlength{\tabcolsep}{2pt}
\renewcommand{\arraystretch}{1.07}
\begin{tabular*}{\linewidth}{@{\extracolsep{\fill}}lrrrrrr@{}}
\toprule
 & \multicolumn{3}{c}{\textbf{MSE $\downarrow$}} & \multicolumn{3}{c}{\textbf{MAE $\downarrow$}} \\
\cmidrule(lr){2-4}\cmidrule(l){5-7}
Model & Base & +Adapter & $\Delta$ (\%) & Base & +Adapter & $\Delta$ (\%) \\
\midrule
DLinear & $0.2270\!\pm\!0.0003$ & $0.2107\!\pm\!0.0000$ & $-7.15$ & $0.3684\!\pm\!0.0004$ & $0.3536\!\pm\!0.0000$ & $-4.03$ \\
xPatch & $0.1481\!\pm\!0.0002$ & $0.1445\!\pm\!0.0001$ & $-2.41$ & $0.2861\!\pm\!0.0002$ & $0.2836\!\pm\!0.0002$ & $-0.89$ \\
PatchTST & $0.1505\!\pm\!0.0006$ & $0.1432\!\pm\!0.0006$ & $-4.84$ & $0.2898\!\pm\!0.0008$ & $0.2836\!\pm\!0.0008$ & $-2.16$ \\
Timer & $0.2039\!\pm\!0.0000$ & $0.1859\!\pm\!0.0000$ & $-8.83$ & $0.3386\!\pm\!0.0000$ & $0.3263\!\pm\!0.0000$ & $-3.63$ \\
TimeMoE & $0.1561\!\pm\!0.0000$ & $0.1482\!\pm\!0.0000$ & $-5.09$ & $0.2917\!\pm\!0.0000$ & $0.2858\!\pm\!0.0000$ & $-2.03$ \\
\midrule
iTransformer & $0.1362\!\pm\!0.0010$ & $0.1328\!\pm\!0.0010$ & $-2.49$ & $0.2730\!\pm\!0.0012$ & $0.2702\!\pm\!0.0011$ & $-1.02$ \\
TQNet & $0.1503\!\pm\!0.0030$ & $0.1466\!\pm\!0.0030$ & $-2.41$ & $0.2890\!\pm\!0.0035$ & $0.2864\!\pm\!0.0034$ & $-0.92$ \\
DUET & $0.1779\!\pm\!0.0381$ & $0.1739\!\pm\!0.0381$ & $-2.29$ & $0.3180\!\pm\!0.0392$ & $0.3147\!\pm\!0.0388$ & $-1.02$ \\
Moirai & $0.1685\!\pm\!0.0000$ & $0.1502\!\pm\!0.0000$ & $-10.86$ & $0.2998\!\pm\!0.0000$ & $0.2875\!\pm\!0.0000$ & $-4.09$ \\
Timer-XL & $0.1266\!\pm\!0.0004$ & $0.1235\!\pm\!0.0001$ & $-2.47$ & $0.2615\!\pm\!0.0004$ & $0.2589\!\pm\!0.0001$ & $-1.00$ \\
\midrule
Corrformer & $0.1237\!\pm\!0.0008$ & $0.1210\!\pm\!0.0007$ & $-2.21$ & $0.2593\!\pm\!0.0008$ & $0.2574\!\pm\!0.0007$ & $-0.74$ \\
S$^2$Transformer & $0.1357\!\pm\!0.0008$ & $0.1346\!\pm\!0.0007$ & $-0.84$ & $0.2785\!\pm\!0.0011$ & $0.2772\!\pm\!0.0010$ & $-0.45$ \\
CDPNet & $0.1156\!\pm\!0.0025$ & $0.1148\!\pm\!0.0022$ & $-0.69$ & $0.2557\!\pm\!0.0011$ & $0.2548\!\pm\!0.0008$ & $-0.35$ \\
EasyST & $0.1371\!\pm\!0.0001$ & $0.1349\!\pm\!0.0000$ & $-1.56$ & $0.2766\!\pm\!0.0001$ & $0.2747\!\pm\!0.0001$ & $-0.70$ \\
STELLA & $0.1754\!\pm\!0.0060$ & $0.1685\!\pm\!0.0052$ & $-3.92$ & $0.3211\!\pm\!0.0061$ & $0.3146\!\pm\!0.0054$ & $-2.02$ \\
\midrule
HiSTGNN & $0.1231\!\pm\!0.0093$ & -- & -- & $0.2622\!\pm\!0.0090$ & -- & -- \\
\bottomrule
\end{tabular*}
\endgroup
\end{table}
\begin{table}[!htbp]
\centering
\caption{\textbf{Europe --- Wind speed.} MSE and MAE (mean $\pm$ sample standard deviation). $\Delta$ (\%) is signed relative change; negative is better.}
\label{tab:europe_ws}
\begingroup
\fontsize{8}{9}\selectfont
\setlength{\tabcolsep}{2pt}
\renewcommand{\arraystretch}{1.07}
\begin{tabular*}{\linewidth}{@{\extracolsep{\fill}}lrrrrrr@{}}
\toprule
 & \multicolumn{3}{c}{\textbf{MSE $\downarrow$}} & \multicolumn{3}{c}{\textbf{MAE $\downarrow$}} \\
\cmidrule(lr){2-4}\cmidrule(l){5-7}
Model & Base & +Adapter & $\Delta$ (\%) & Base & +Adapter & $\Delta$ (\%) \\
\midrule
DLinear & $0.7531\!\pm\!0.0004$ & $0.6782\!\pm\!0.0000$ & $-9.94$ & $0.6265\!\pm\!0.0014$ & $0.6085\!\pm\!0.0000$ & $-2.87$ \\
xPatch & $0.6890\!\pm\!0.0004$ & $0.6318\!\pm\!0.0002$ & $-8.29$ & $0.5836\!\pm\!0.0002$ & $0.5780\!\pm\!0.0002$ & $-0.96$ \\
PatchTST & $0.6775\!\pm\!0.0008$ & $0.6272\!\pm\!0.0001$ & $-7.42$ & $0.5830\!\pm\!0.0002$ & $0.5762\!\pm\!0.0001$ & $-1.18$ \\
Timer & $0.7625\!\pm\!0.0000$ & $0.6617\!\pm\!0.0000$ & $-13.23$ & $0.6238\!\pm\!0.0000$ & $0.5979\!\pm\!0.0000$ & $-4.16$ \\
TimeMoE & $0.7422\!\pm\!0.0000$ & $0.6554\!\pm\!0.0000$ & $-11.68$ & $0.6048\!\pm\!0.0000$ & $0.5905\!\pm\!0.0000$ & $-2.35$ \\
\midrule
iTransformer & $0.6455\!\pm\!0.0006$ & $0.6171\!\pm\!0.0007$ & $-4.40$ & $0.5682\!\pm\!0.0008$ & $0.5680\!\pm\!0.0005$ & $-0.05$ \\
TQNet & $0.6807\!\pm\!0.0035$ & $0.6294\!\pm\!0.0005$ & $-7.53$ & $0.5838\!\pm\!0.0025$ & $0.5799\!\pm\!0.0022$ & $-0.66$ \\
DUET & $0.7174\!\pm\!0.0349$ & $0.6588\!\pm\!0.0256$ & $-8.17$ & $0.5938\!\pm\!0.0182$ & $0.5901\!\pm\!0.0164$ & $-0.63$ \\
Moirai & $0.8183\!\pm\!0.0000$ & $0.6647\!\pm\!0.0000$ & $-18.78$ & $0.6253\!\pm\!0.0000$ & $0.5948\!\pm\!0.0000$ & $-4.88$ \\
Timer-XL & $0.6281\!\pm\!0.0003$ & $0.6086\!\pm\!0.0004$ & $-3.11$ & $0.5599\!\pm\!0.0015$ & $0.5601\!\pm\!0.0002$ & $+0.03$ \\
\midrule
Corrformer & $0.6263\!\pm\!0.0019$ & $0.6035\!\pm\!0.0011$ & $-3.64$ & $0.5598\!\pm\!0.0011$ & $0.5594\!\pm\!0.0006$ & $-0.08$ \\
S$^2$Transformer & $0.6162\!\pm\!0.0006$ & $0.6134\!\pm\!0.0005$ & $-0.45$ & $0.5661\!\pm\!0.0004$ & $0.5657\!\pm\!0.0003$ & $-0.07$ \\
CDPNet & $0.5875\!\pm\!0.0007$ & $0.5722\!\pm\!0.0032$ & $-2.62$ & $0.5358\!\pm\!0.0018$ & $0.5430\!\pm\!0.0023$ & $+1.34$ \\
EasyST & $0.5909\!\pm\!0.0004$ & $0.5899\!\pm\!0.0005$ & $-0.16$ & $0.5539\!\pm\!0.0006$ & $0.5518\!\pm\!0.0003$ & $-0.37$ \\
STELLA & $0.7013\!\pm\!0.0031$ & $0.6557\!\pm\!0.0037$ & $-6.51$ & $0.5997\!\pm\!0.0050$ & $0.5874\!\pm\!0.0085$ & $-2.06$ \\
\midrule
HiSTGNN & $0.5662\!\pm\!0.0105$ & -- & -- & $0.5264\!\pm\!0.0047$ & -- & -- \\
\bottomrule
\end{tabular*}
\endgroup
\end{table}
\begin{table}[!htbp]
\centering
\caption{\textbf{Europe --- Relative humidity.} MSE and MAE (mean $\pm$ sample standard deviation). $\Delta$ (\%) is signed relative change; negative is better.}
\label{tab:europe_rh}
\begingroup
\fontsize{8}{9}\selectfont
\setlength{\tabcolsep}{2pt}
\renewcommand{\arraystretch}{1.07}
\begin{tabular*}{\linewidth}{@{\extracolsep{\fill}}lrrrrrr@{}}
\toprule
 & \multicolumn{3}{c}{\textbf{MSE $\downarrow$}} & \multicolumn{3}{c}{\textbf{MAE $\downarrow$}} \\
\cmidrule(lr){2-4}\cmidrule(l){5-7}
Model & Base & +Adapter & $\Delta$ (\%) & Base & +Adapter & $\Delta$ (\%) \\
\midrule
DLinear & $0.7719\!\pm\!0.0006$ & $0.7191\!\pm\!0.0000$ & $-6.85$ & $0.6820\!\pm\!0.0019$ & $0.6794\!\pm\!0.0000$ & $-0.39$ \\
xPatch & $0.5427\!\pm\!0.0001$ & $0.5119\!\pm\!0.0002$ & $-5.68$ & $0.5439\!\pm\!0.0003$ & $0.5416\!\pm\!0.0003$ & $-0.43$ \\
PatchTST & $0.5371\!\pm\!0.0015$ & $0.5131\!\pm\!0.0016$ & $-4.46$ & $0.5502\!\pm\!0.0015$ & $0.5499\!\pm\!0.0011$ & $-0.05$ \\
Timer & $0.7665\!\pm\!0.0000$ & $0.6671\!\pm\!0.0000$ & $-12.98$ & $0.6468\!\pm\!0.0000$ & $0.6440\!\pm\!0.0000$ & $-0.43$ \\
TimeMoE & $0.5974\!\pm\!0.0000$ & $0.5476\!\pm\!0.0000$ & $-8.33$ & $0.5686\!\pm\!0.0000$ & $0.5617\!\pm\!0.0000$ & $-1.22$ \\
\midrule
iTransformer & $0.4993\!\pm\!0.0008$ & $0.4768\!\pm\!0.0013$ & $-4.51$ & $0.5237\!\pm\!0.0007$ & $0.5304\!\pm\!0.0011$ & $+1.28$ \\
TQNet & $0.5341\!\pm\!0.0081$ & $0.5099\!\pm\!0.0063$ & $-4.53$ & $0.5439\!\pm\!0.0025$ & $0.5410\!\pm\!0.0039$ & $-0.54$ \\
DUET & $0.6405\!\pm\!0.1267$ & $0.5999\!\pm\!0.1177$ & $-6.34$ & $0.5975\!\pm\!0.0695$ & $0.6044\!\pm\!0.0727$ & $+1.15$ \\
Moirai & $0.6382\!\pm\!0.0000$ & $0.5370\!\pm\!0.0000$ & $-15.86$ & $0.5755\!\pm\!0.0000$ & $0.5647\!\pm\!0.0000$ & $-1.88$ \\
Timer-XL & $0.4786\!\pm\!0.0017$ & $0.4610\!\pm\!0.0004$ & $-3.67$ & $0.5116\!\pm\!0.0011$ & $0.5102\!\pm\!0.0003$ & $-0.26$ \\
\midrule
Corrformer & $0.4681\!\pm\!0.0009$ & $0.4494\!\pm\!0.0008$ & $-4.00$ & $0.5017\!\pm\!0.0005$ & $0.5079\!\pm\!0.0004$ & $+1.24$ \\
S$^2$Transformer & $0.4888\!\pm\!0.0007$ & $0.4860\!\pm\!0.0008$ & $-0.58$ & $0.5317\!\pm\!0.0006$ & $0.5314\!\pm\!0.0005$ & $-0.07$ \\
CDPNet & $0.4530\!\pm\!0.0008$ & $0.4443\!\pm\!0.0025$ & $-1.93$ & $0.5019\!\pm\!0.0013$ & $0.5009\!\pm\!0.0009$ & $-0.18$ \\
EasyST & $0.4943\!\pm\!0.0005$ & $0.4889\!\pm\!0.0006$ & $-1.09$ & $0.5365\!\pm\!0.0002$ & $0.5341\!\pm\!0.0004$ & $-0.44$ \\
STELLA & $0.5630\!\pm\!0.0046$ & $0.5310\!\pm\!0.0047$ & $-5.69$ & $0.5649\!\pm\!0.0057$ & $0.5643\!\pm\!0.0060$ & $-0.11$ \\
\midrule
HiSTGNN & $0.4548\!\pm\!0.0210$ & -- & -- & $0.5000\!\pm\!0.0092$ & -- & -- \\
\bottomrule
\end{tabular*}
\endgroup
\end{table}
\begin{table}[!htbp]
\centering
\caption{\textbf{Europe --- Station pressure.} MSE and MAE (mean $\pm$ sample standard deviation). $\Delta$ (\%) is signed relative change; negative is better.}
\label{tab:europe_p}
\begingroup
\fontsize{8}{9}\selectfont
\setlength{\tabcolsep}{2pt}
\renewcommand{\arraystretch}{1.07}
\begin{tabular*}{\linewidth}{@{\extracolsep{\fill}}lrrrrrr@{}}
\toprule
 & \multicolumn{3}{c}{\textbf{MSE $\downarrow$}} & \multicolumn{3}{c}{\textbf{MAE $\downarrow$}} \\
\cmidrule(lr){2-4}\cmidrule(l){5-7}
Model & Base & +Adapter & $\Delta$ (\%) & Base & +Adapter & $\Delta$ (\%) \\
\midrule
DLinear & $0.0404\!\pm\!0.0004$ & $0.0268\!\pm\!0.0000$ & $-33.59$ & $0.1505\!\pm\!0.0017$ & $0.1128\!\pm\!0.0000$ & $-25.07$ \\
xPatch & $0.0238\!\pm\!0.0001$ & $0.0236\!\pm\!0.0001$ & $-0.64$ & $0.1023\!\pm\!0.0003$ & $0.1022\!\pm\!0.0003$ & $-0.06$ \\
PatchTST & $0.0236\!\pm\!0.0001$ & $0.0229\!\pm\!0.0001$ & $-2.99$ & $0.1017\!\pm\!0.0003$ & $0.1003\!\pm\!0.0003$ & $-1.34$ \\
Timer & $0.0240\!\pm\!0.0000$ & $0.0232\!\pm\!0.0000$ & $-3.04$ & $0.1016\!\pm\!0.0000$ & $0.1001\!\pm\!0.0000$ & $-1.48$ \\
TimeMoE & $0.0234\!\pm\!0.0000$ & $0.0227\!\pm\!0.0000$ & $-3.23$ & $0.0986\!\pm\!0.0000$ & $0.0972\!\pm\!0.0000$ & $-1.47$ \\
\midrule
iTransformer & $0.0214\!\pm\!0.0000$ & $0.0212\!\pm\!0.0000$ & $-0.73$ & $0.0937\!\pm\!0.0000$ & $0.0936\!\pm\!0.0000$ & $-0.10$ \\
TQNet & $0.0236\!\pm\!0.0003$ & $0.0235\!\pm\!0.0002$ & $-0.52$ & $0.1013\!\pm\!0.0010$ & $0.1013\!\pm\!0.0009$ & $0.00$ \\
DUET & $0.0262\!\pm\!0.0015$ & $0.0259\!\pm\!0.0014$ & $-0.94$ & $0.1089\!\pm\!0.0053$ & $0.1085\!\pm\!0.0052$ & $-0.29$ \\
Moirai & $0.0245\!\pm\!0.0000$ & $0.0234\!\pm\!0.0000$ & $-4.55$ & $0.0996\!\pm\!0.0000$ & $0.0974\!\pm\!0.0000$ & $-2.17$ \\
Timer-XL & $0.0196\!\pm\!0.0000$ & $0.0195\!\pm\!0.0000$ & $-0.85$ & $0.0882\!\pm\!0.0001$ & $0.0880\!\pm\!0.0001$ & $-0.26$ \\
\midrule
Corrformer & $0.0216\!\pm\!0.0003$ & $0.0212\!\pm\!0.0003$ & $-1.64$ & $0.0941\!\pm\!0.0008$ & $0.0936\!\pm\!0.0008$ & $-0.54$ \\
S$^2$Transformer & $0.0921\!\pm\!0.0031$ & $0.0670\!\pm\!0.0017$ & $-27.23$ & $0.1414\!\pm\!0.0013$ & $0.1405\!\pm\!0.0018$ & $-0.62$ \\
CDPNet & $0.0199\!\pm\!0.0006$ & $0.0195\!\pm\!0.0004$ & $-1.84$ & $0.0919\!\pm\!0.0031$ & $0.0900\!\pm\!0.0018$ & $-2.13$ \\
EasyST & $0.0209\!\pm\!0.0001$ & $0.0206\!\pm\!0.0001$ & $-1.42$ & $0.0949\!\pm\!0.0004$ & $0.0936\!\pm\!0.0005$ & $-1.42$ \\
STELLA & $0.0530\!\pm\!0.0055$ & $0.0528\!\pm\!0.0055$ & $-0.45$ & $0.1743\!\pm\!0.0105$ & $0.1737\!\pm\!0.0105$ & $-0.33$ \\
\midrule
HiSTGNN & $0.0192\!\pm\!0.0008$ & -- & -- & $0.0919\!\pm\!0.0029$ & -- & -- \\
\bottomrule
\end{tabular*}
\endgroup
\end{table}
\begin{table}[!htbp]
\centering
\caption{\textbf{Global --- Air temperature.} MSE and MAE (mean $\pm$ sample standard deviation). $\Delta$ (\%) is signed relative change; negative is better.}
\label{tab:global_t}
\begingroup
\fontsize{8}{9}\selectfont
\setlength{\tabcolsep}{2pt}
\renewcommand{\arraystretch}{1.07}
\begin{tabular*}{\linewidth}{@{\extracolsep{\fill}}lrrrrrr@{}}
\toprule
 & \multicolumn{3}{c}{\textbf{MSE $\downarrow$}} & \multicolumn{3}{c}{\textbf{MAE $\downarrow$}} \\
\cmidrule(lr){2-4}\cmidrule(l){5-7}
Model & Base & +Adapter & $\Delta$ (\%) & Base & +Adapter & $\Delta$ (\%) \\
\midrule
DLinear & $0.1837\!\pm\!0.0008$ & $0.1685\!\pm\!0.0011$ & $-8.25$ & $0.3285\!\pm\!0.0009$ & $0.3124\!\pm\!0.0012$ & $-4.88$ \\
xPatch & $0.1186\!\pm\!0.0010$ & $0.1160\!\pm\!0.0011$ & $-2.25$ & $0.2507\!\pm\!0.0015$ & $0.2495\!\pm\!0.0015$ & $-0.46$ \\
PatchTST & $0.1188\!\pm\!0.0000$ & $0.1139\!\pm\!0.0001$ & $-4.11$ & $0.2514\!\pm\!0.0001$ & $0.2478\!\pm\!0.0001$ & $-1.45$ \\
Timer & $0.1628\!\pm\!0.0000$ & $0.1500\!\pm\!0.0000$ & $-7.88$ & $0.2979\!\pm\!0.0000$ & $0.2895\!\pm\!0.0000$ & $-2.80$ \\
TimeMoE & $0.1258\!\pm\!0.0000$ & $0.1208\!\pm\!0.0000$ & $-3.95$ & $0.2562\!\pm\!0.0000$ & $0.2533\!\pm\!0.0000$ & $-1.12$ \\
\midrule
iTransformer & $0.1019\!\pm\!0.0013$ & $0.0999\!\pm\!0.0012$ & $-1.89$ & $0.2297\!\pm\!0.0017$ & $0.2289\!\pm\!0.0017$ & $-0.36$ \\
TQNet & $0.1176\!\pm\!0.0017$ & $0.1149\!\pm\!0.0019$ & $-2.25$ & $0.2493\!\pm\!0.0024$ & $0.2481\!\pm\!0.0028$ & $-0.47$ \\
DUET & $0.1223\!\pm\!0.0022$ & $0.1197\!\pm\!0.0019$ & $-2.09$ & $0.2559\!\pm\!0.0027$ & $0.2548\!\pm\!0.0026$ & $-0.45$ \\
Moirai & $0.1398\!\pm\!0.0000$ & $0.1262\!\pm\!0.0000$ & $-9.74$ & $0.2661\!\pm\!0.0000$ & $0.2576\!\pm\!0.0000$ & $-3.19$ \\
Timer-XL & $0.0983\!\pm\!0.0001$ & $0.0961\!\pm\!0.0001$ & $-2.23$ & $0.2250\!\pm\!0.0001$ & $0.2236\!\pm\!0.0001$ & $-0.60$ \\
\midrule
Corrformer & $0.0986\!\pm\!0.0003$ & $0.0969\!\pm\!0.0004$ & $-1.66$ & $0.2272\!\pm\!0.0005$ & $0.2267\!\pm\!0.0006$ & $-0.22$ \\
S$^2$Transformer & $0.0991\!\pm\!0.0008$ & $0.0986\!\pm\!0.0007$ & $-0.46$ & $0.2337\!\pm\!0.0010$ & $0.2333\!\pm\!0.0010$ & $-0.21$ \\
CDPNet & $0.0863\!\pm\!0.0007$ & $0.0857\!\pm\!0.0008$ & $-0.73$ & $0.2183\!\pm\!0.0014$ & $0.2174\!\pm\!0.0014$ & $-0.39$ \\
EasyST & $0.1072\!\pm\!0.0004$ & $0.1067\!\pm\!0.0004$ & $-0.44$ & $0.2393\!\pm\!0.0007$ & $0.2387\!\pm\!0.0006$ & $-0.26$ \\
STELLA & $0.1902\!\pm\!0.0519$ & $0.1738\!\pm\!0.0401$ & $-8.63$ & $0.3342\!\pm\!0.0519$ & $0.3192\!\pm\!0.0420$ & $-4.49$ \\
\midrule
HiSTGNN & $0.0949\!\pm\!0.0029$ & -- & -- & $0.2267\!\pm\!0.0037$ & -- & -- \\
\bottomrule
\end{tabular*}
\endgroup
\end{table}
\begin{table}[!htbp]
\centering
\caption{\textbf{Global --- Wind speed.} MSE and MAE (mean $\pm$ sample standard deviation). $\Delta$ (\%) is signed relative change; negative is better.}
\label{tab:global_ws}
\begingroup
\fontsize{8}{9}\selectfont
\setlength{\tabcolsep}{2pt}
\renewcommand{\arraystretch}{1.07}
\begin{tabular*}{\linewidth}{@{\extracolsep{\fill}}lrrrrrr@{}}
\toprule
 & \multicolumn{3}{c}{\textbf{MSE $\downarrow$}} & \multicolumn{3}{c}{\textbf{MAE $\downarrow$}} \\
\cmidrule(lr){2-4}\cmidrule(l){5-7}
Model & Base & +Adapter & $\Delta$ (\%) & Base & +Adapter & $\Delta$ (\%) \\
\midrule
DLinear & $0.8466\!\pm\!0.0053$ & $0.7770\!\pm\!0.0010$ & $-8.22$ & $0.6877\!\pm\!0.0024$ & $0.6675\!\pm\!0.0006$ & $-2.94$ \\
xPatch & $0.7833\!\pm\!0.0041$ & $0.7216\!\pm\!0.0003$ & $-7.87$ & $0.6473\!\pm\!0.0013$ & $0.6348\!\pm\!0.0004$ & $-1.94$ \\
PatchTST & $0.7767\!\pm\!0.0005$ & $0.7179\!\pm\!0.0002$ & $-7.57$ & $0.6473\!\pm\!0.0002$ & $0.6333\!\pm\!0.0001$ & $-2.17$ \\
Timer & $0.8832\!\pm\!0.0000$ & $0.7617\!\pm\!0.0000$ & $-13.76$ & $0.6968\!\pm\!0.0000$ & $0.6576\!\pm\!0.0000$ & $-5.62$ \\
TimeMoE & $0.8533\!\pm\!0.0000$ & $0.7477\!\pm\!0.0000$ & $-12.38$ & $0.6747\!\pm\!0.0000$ & $0.6476\!\pm\!0.0000$ & $-4.01$ \\
\midrule
iTransformer & $0.7134\!\pm\!0.0037$ & $0.6843\!\pm\!0.0019$ & $-4.08$ & $0.6177\!\pm\!0.0018$ & $0.6133\!\pm\!0.0013$ & $-0.71$ \\
TQNet & $0.7687\!\pm\!0.0163$ & $0.7181\!\pm\!0.0036$ & $-6.58$ & $0.6431\!\pm\!0.0086$ & $0.6343\!\pm\!0.0027$ & $-1.37$ \\
DUET & $0.7844\!\pm\!0.0109$ & $0.7264\!\pm\!0.0044$ & $-7.40$ & $0.6451\!\pm\!0.0049$ & $0.6474\!\pm\!0.0026$ & $+0.36$ \\
Moirai & $0.9503\!\pm\!0.0000$ & $0.7581\!\pm\!0.0000$ & $-20.22$ & $0.7032\!\pm\!0.0000$ & $0.6528\!\pm\!0.0000$ & $-7.16$ \\
Timer-XL & $0.7031\!\pm\!0.0001$ & $0.6793\!\pm\!0.0002$ & $-3.38$ & $0.6134\!\pm\!0.0010$ & $0.6104\!\pm\!0.0001$ & $-0.48$ \\
\midrule
Corrformer & $0.7069\!\pm\!0.0007$ & $0.6778\!\pm\!0.0003$ & $-4.11$ & $0.6166\!\pm\!0.0009$ & $0.6096\!\pm\!0.0004$ & $-1.13$ \\
S$^2$Transformer & $0.6816\!\pm\!0.0005$ & $0.6809\!\pm\!0.0005$ & $-0.10$ & $0.6127\!\pm\!0.0002$ & $0.6123\!\pm\!0.0001$ & $-0.06$ \\
CDPNet & $0.6526\!\pm\!0.0043$ & $0.6472\!\pm\!0.0059$ & $-0.83$ & $0.5899\!\pm\!0.0035$ & $0.5941\!\pm\!0.0026$ & $+0.70$ \\
EasyST & $0.6597\!\pm\!0.0004$ & $0.6572\!\pm\!0.0003$ & $-0.38$ & $0.6038\!\pm\!0.0002$ & $0.6023\!\pm\!0.0001$ & $-0.25$ \\
STELLA & $0.8433\!\pm\!0.0545$ & $0.7585\!\pm\!0.0257$ & $-10.05$ & $0.6836\!\pm\!0.0332$ & $0.6527\!\pm\!0.0183$ & $-4.53$ \\
\midrule
HiSTGNN & $0.6423\!\pm\!0.0059$ & -- & -- & $0.5815\!\pm\!0.0013$ & -- & -- \\
\bottomrule
\end{tabular*}
\endgroup
\end{table}
\begin{table}[!htbp]
\centering
\caption{\textbf{Global --- Relative humidity.} MSE and MAE (mean $\pm$ sample standard deviation). $\Delta$ (\%) is signed relative change; negative is better.}
\label{tab:global_rh}
\begingroup
\fontsize{8}{9}\selectfont
\setlength{\tabcolsep}{2pt}
\renewcommand{\arraystretch}{1.07}
\begin{tabular*}{\linewidth}{@{\extracolsep{\fill}}lrrrrrr@{}}
\toprule
 & \multicolumn{3}{c}{\textbf{MSE $\downarrow$}} & \multicolumn{3}{c}{\textbf{MAE $\downarrow$}} \\
\cmidrule(lr){2-4}\cmidrule(l){5-7}
Model & Base & +Adapter & $\Delta$ (\%) & Base & +Adapter & $\Delta$ (\%) \\
\midrule
DLinear & $0.7290\!\pm\!0.0071$ & $0.6969\!\pm\!0.0040$ & $-4.41$ & $0.6677\!\pm\!0.0031$ & $0.6673\!\pm\!0.0023$ & $-0.05$ \\
xPatch & $0.5100\!\pm\!0.0042$ & $0.4878\!\pm\!0.0035$ & $-4.36$ & $0.5264\!\pm\!0.0023$ & $0.5214\!\pm\!0.0024$ & $-0.95$ \\
PatchTST & $0.4980\!\pm\!0.0006$ & $0.4829\!\pm\!0.0008$ & $-3.03$ & $0.5285\!\pm\!0.0009$ & $0.5261\!\pm\!0.0006$ & $-0.46$ \\
Timer & $0.7494\!\pm\!0.0000$ & $0.6601\!\pm\!0.0000$ & $-11.92$ & $0.6430\!\pm\!0.0000$ & $0.6371\!\pm\!0.0000$ & $-0.91$ \\
TimeMoE & $0.5753\!\pm\!0.0000$ & $0.5332\!\pm\!0.0000$ & $-7.32$ & $0.5577\!\pm\!0.0000$ & $0.5611\!\pm\!0.0000$ & $+0.61$ \\
\midrule
iTransformer & $0.4577\!\pm\!0.0031$ & $0.4435\!\pm\!0.0025$ & $-3.11$ & $0.4972\!\pm\!0.0018$ & $0.4903\!\pm\!0.0017$ & $-1.38$ \\
TQNet & $0.4930\!\pm\!0.0090$ & $0.4771\!\pm\!0.0090$ & $-3.23$ & $0.5211\!\pm\!0.0064$ & $0.5246\!\pm\!0.0073$ & $+0.66$ \\
DUET & $0.5236\!\pm\!0.0085$ & $0.4961\!\pm\!0.0068$ & $-5.25$ & $0.5332\!\pm\!0.0053$ & $0.5308\!\pm\!0.0050$ & $-0.45$ \\
Moirai & $0.6141\!\pm\!0.0000$ & $0.5248\!\pm\!0.0000$ & $-14.55$ & $0.5638\!\pm\!0.0000$ & $0.5555\!\pm\!0.0000$ & $-1.47$ \\
Timer-XL & $0.4505\!\pm\!0.0007$ & $0.4371\!\pm\!0.0006$ & $-2.97$ & $0.4938\!\pm\!0.0006$ & $0.4882\!\pm\!0.0004$ & $-1.13$ \\
\midrule
Corrformer & $0.4496\!\pm\!0.0019$ & $0.4367\!\pm\!0.0010$ & $-2.87$ & $0.4924\!\pm\!0.0010$ & $0.4980\!\pm\!0.0008$ & $+1.14$ \\
S$^2$Transformer & $0.4467\!\pm\!0.0011$ & $0.4462\!\pm\!0.0011$ & $-0.11$ & $0.5065\!\pm\!0.0005$ & $0.5059\!\pm\!0.0006$ & $-0.13$ \\
CDPNet & $0.4382\!\pm\!0.0041$ & $0.4349\!\pm\!0.0040$ & $-0.76$ & $0.4965\!\pm\!0.0023$ & $0.4960\!\pm\!0.0023$ & $-0.10$ \\
EasyST & $0.4548\!\pm\!0.0004$ & $0.4523\!\pm\!0.0003$ & $-0.56$ & $0.5110\!\pm\!0.0005$ & $0.5089\!\pm\!0.0004$ & $-0.41$ \\
STELLA & $0.5912\!\pm\!0.0619$ & $0.5540\!\pm\!0.0455$ & $-6.29$ & $0.5847\!\pm\!0.0390$ & $0.5781\!\pm\!0.0278$ & $-1.12$ \\
\midrule
HiSTGNN & $0.4291\!\pm\!0.0053$ & -- & -- & $0.4855\!\pm\!0.0052$ & -- & -- \\
\bottomrule
\end{tabular*}
\endgroup
\end{table}
\begin{table}[!htbp]
\centering
\caption{\textbf{Global --- Station pressure.} MSE and MAE (mean $\pm$ sample standard deviation). $\Delta$ (\%) is signed relative change; negative is better.}
\label{tab:global_p}
\begingroup
\fontsize{8}{9}\selectfont
\setlength{\tabcolsep}{2pt}
\renewcommand{\arraystretch}{1.07}
\begin{tabular*}{\linewidth}{@{\extracolsep{\fill}}lrrrrrr@{}}
\toprule
 & \multicolumn{3}{c}{\textbf{MSE $\downarrow$}} & \multicolumn{3}{c}{\textbf{MAE $\downarrow$}} \\
\cmidrule(lr){2-4}\cmidrule(l){5-7}
Model & Base & +Adapter & $\Delta$ (\%) & Base & +Adapter & $\Delta$ (\%) \\
\midrule
DLinear & $0.0438\!\pm\!0.0013$ & $0.0263\!\pm\!0.0000$ & $-39.83$ & $0.1603\!\pm\!0.0019$ & $0.1163\!\pm\!0.0001$ & $-27.50$ \\
xPatch & $0.0230\!\pm\!0.0004$ & $0.0229\!\pm\!0.0004$ & $-0.59$ & $0.1043\!\pm\!0.0012$ & $0.1043\!\pm\!0.0012$ & $-0.08$ \\
PatchTST & $0.0224\!\pm\!0.0001$ & $0.0218\!\pm\!0.0001$ & $-2.71$ & $0.1026\!\pm\!0.0004$ & $0.1015\!\pm\!0.0004$ & $-1.05$ \\
Timer & $0.0232\!\pm\!0.0000$ & $0.0225\!\pm\!0.0000$ & $-2.85$ & $0.1046\!\pm\!0.0000$ & $0.1033\!\pm\!0.0000$ & $-1.22$ \\
TimeMoE & $0.0227\!\pm\!0.0000$ & $0.0220\!\pm\!0.0000$ & $-3.06$ & $0.1021\!\pm\!0.0000$ & $0.1009\!\pm\!0.0000$ & $-1.16$ \\
\midrule
iTransformer & $0.0200\!\pm\!0.0003$ & $0.0199\!\pm\!0.0003$ & $-0.56$ & $0.0943\!\pm\!0.0011$ & $0.0933\!\pm\!0.0011$ & $-1.00$ \\
TQNet & $0.0230\!\pm\!0.0005$ & $0.0229\!\pm\!0.0004$ & $-0.47$ & $0.1040\!\pm\!0.0015$ & $0.1040\!\pm\!0.0014$ & $-0.04$ \\
DUET & $0.0258\!\pm\!0.0036$ & $0.0256\!\pm\!0.0035$ & $-0.74$ & $0.1113\!\pm\!0.0081$ & $0.1112\!\pm\!0.0083$ & $-0.07$ \\
Moirai & $0.0262\!\pm\!0.0000$ & $0.0249\!\pm\!0.0000$ & $-5.09$ & $0.1074\!\pm\!0.0000$ & $0.1053\!\pm\!0.0000$ & $-1.95$ \\
Timer-XL & $0.0190\!\pm\!0.0000$ & $0.0189\!\pm\!0.0000$ & $-0.43$ & $0.0911\!\pm\!0.0001$ & $0.0901\!\pm\!0.0000$ & $-1.05$ \\
\midrule
Corrformer & $0.0205\!\pm\!0.0003$ & $0.0202\!\pm\!0.0003$ & $-1.30$ & $0.0964\!\pm\!0.0009$ & $0.0961\!\pm\!0.0008$ & $-0.35$ \\
S$^2$Transformer & $0.0358\!\pm\!0.0006$ & $0.0334\!\pm\!0.0005$ & $-6.67$ & $0.1086\!\pm\!0.0007$ & $0.1008\!\pm\!0.0007$ & $-7.11$ \\
CDPNet & $0.0168\!\pm\!0.0003$ & $0.0165\!\pm\!0.0002$ & $-1.86$ & $0.0888\!\pm\!0.0010$ & $0.0879\!\pm\!0.0009$ & $-0.98$ \\
EasyST & $0.0186\!\pm\!0.0001$ & $0.0185\!\pm\!0.0001$ & $-0.54$ & $0.0933\!\pm\!0.0006$ & $0.0926\!\pm\!0.0004$ & $-0.67$ \\
STELLA & $0.0894\!\pm\!0.0453$ & $0.0862\!\pm\!0.0424$ & $-3.60$ & $0.2249\!\pm\!0.0669$ & $0.2212\!\pm\!0.0632$ & $-1.60$ \\
\midrule
HiSTGNN & $0.0179\!\pm\!0.0002$ & -- & -- & $0.0927\!\pm\!0.0011$ & -- & -- \\
\bottomrule
\end{tabular*}
\endgroup
\end{table}

\subsection{Individual Models within Each Paradigm}
\label{app:model_results}
\begin{figure}[!htbp]
\centering
\includegraphics[width=\linewidth]{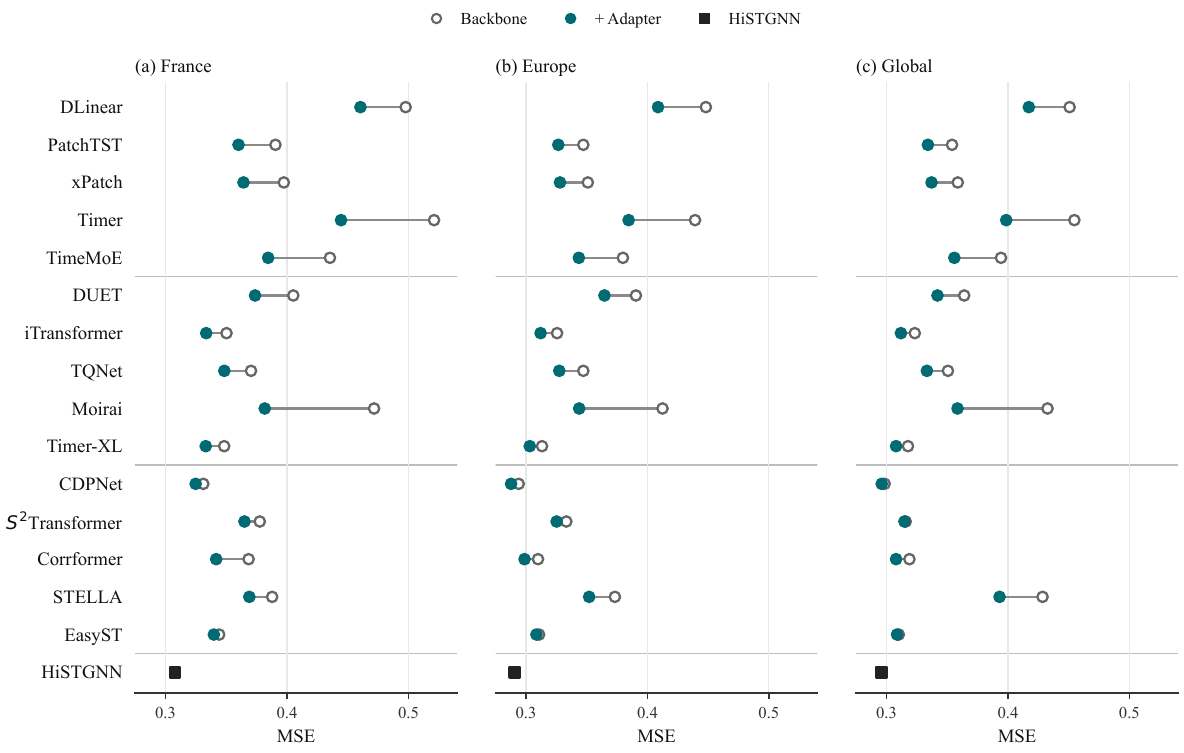}
\caption{Individual-model counterpart of Figure~\ref{fig:paradigm_results}. Open circles denote native predictions, filled circles denote adapted predictions, and squares mark the unadapted HiSTGNN. Thin rules separate SSSV, SSMV, MSSV, and MSMV; lower MSE is better.}
\label{fig:model_results}
\end{figure}

Figure~\ref{fig:model_results} expands Figure~\ref{fig:paradigm_results} into all 16 backbones. The same model order and horizontal scale are used across datasets. Each connected pair compares a frozen backbone with its full adapter; HiSTGNN appears only as the native MSMV reference. Even within the same station--variable modeling paradigm, backbones benefit from interaction completion to different degrees. For example, among SSMV models on France, MSE decreases by 19.08\% for Moirai but 4.36\% for Timer-XL. Thus, the modeling paradigm alone does not determine the magnitude of the improvement; the benefit also varies with the backbone.

\subsection{Removing the Calibration Branch}
\label{app:interaction_ablation}
The interaction-only variant fixes $a_{h,v}=b_{h,v}=0$ and refits the same enabled interaction branches as the full adapter, with the backbone frozen. Table~\ref{tab:interaction_ablation} reports MSE for France and Global. Interaction-only correction lowers MSE for all 15 backbones on both datasets, showing that useful gains persist without the explicit calibration term. The full adapter further improves every pair. For example, on Global, Moirai improves from 0.4326 to 0.3766 with interaction-only correction and to 0.3585 with the full adapter. These separately fitted variants establish complementary benefits without treating their improvements as an additive decomposition.

\begin{table}[!htbp]
\centering
\caption{Calibration ablation on France and Global. MSE is reported as mean $\pm$ sample standard deviation across seeds 2024--2026. Lower is better. \textbf{Full} includes calibration and the same interaction branches. HiSTGNN is excluded.}
\label{tab:interaction_ablation}
\begingroup
\small
\setlength{\tabcolsep}{8pt}
\renewcommand{\arraystretch}{1.08}
\begin{tabular*}{\linewidth}{@{\extracolsep{\fill}}lrrr@{}}
\toprule
\multicolumn{4}{c}{\textbf{France}} \\
\midrule
Model & Base & Interaction only & Full \\
\midrule
DLinear & $0.4977\,\pm\,0.0004$ & $0.4800\,\pm\,0.0004$ & $\mathbf{0.4605}\,\pm\,0.0000$ \\
xPatch & $0.3974\,\pm\,0.0004$ & $0.3858\,\pm\,0.0003$ & $\mathbf{0.3642}\,\pm\,0.0002$ \\
PatchTST & $0.3905\,\pm\,0.0004$ & $0.3724\,\pm\,0.0004$ & $\mathbf{0.3601}\,\pm\,0.0003$ \\
Timer & $0.5211\,\pm\,0.0000$ & $0.4704\,\pm\,0.0000$ & $\mathbf{0.4445}\,\pm\,0.0000$ \\
TimeMoE & $0.4354\,\pm\,0.0000$ & $0.4028\,\pm\,0.0000$ & $\mathbf{0.3845}\,\pm\,0.0000$ \\
\midrule
iTransformer & $0.3502\,\pm\,0.0001$ & $0.3423\,\pm\,0.0000$ & $\mathbf{0.3335}\,\pm\,0.0001$ \\
TQNet & $0.3704\,\pm\,0.0004$ & $0.3628\,\pm\,0.0004$ & $\mathbf{0.3484}\,\pm\,0.0003$ \\
DUET & $0.4052\,\pm\,0.0074$ & $0.3949\,\pm\,0.0066$ & $\mathbf{0.3737}\,\pm\,0.0062$ \\
Moirai & $0.4717\,\pm\,0.0000$ & $0.4094\,\pm\,0.0000$ & $\mathbf{0.3817}\,\pm\,0.0000$ \\
Timer-XL & $0.3483\,\pm\,0.0001$ & $0.3401\,\pm\,0.0002$ & $\mathbf{0.3331}\,\pm\,0.0002$ \\
\midrule
Corrformer & $0.3684\,\pm\,0.0031$ & $0.3556\,\pm\,0.0039$ & $\mathbf{0.3417}\,\pm\,0.0043$ \\
S$^2$Transformer & $0.3776\,\pm\,0.0015$ & $0.3743\,\pm\,0.0014$ & $\mathbf{0.3650}\,\pm\,0.0012$ \\
CDPNet & $0.3311\,\pm\,0.0057$ & $0.3293\,\pm\,0.0063$ & $\mathbf{0.3248}\,\pm\,0.0083$ \\
EasyST & $0.3442\,\pm\,0.0009$ & $0.3428\,\pm\,0.0009$ & $\mathbf{0.3398}\,\pm\,0.0008$ \\
STELLA & $0.3878\,\pm\,0.0005$ & $0.3798\,\pm\,0.0007$ & $\mathbf{0.3690}\,\pm\,0.0007$ \\
\midrule
\multicolumn{4}{c}{\textbf{Global}} \\
\midrule
Model & Base & Interaction only & Full \\
\midrule
DLinear & $0.4508\,\pm\,0.0030$ & $0.4369\,\pm\,0.0022$ & $\mathbf{0.4172}\,\pm\,0.0015$ \\
xPatch & $0.3587\,\pm\,0.0024$ & $0.3478\,\pm\,0.0019$ & $\mathbf{0.3371}\,\pm\,0.0012$ \\
PatchTST & $0.3540\,\pm\,0.0001$ & $0.3414\,\pm\,0.0002$ & $\mathbf{0.3341}\,\pm\,0.0002$ \\
Timer & $0.4547\,\pm\,0.0000$ & $0.4154\,\pm\,0.0000$ & $\mathbf{0.3986}\,\pm\,0.0000$ \\
TimeMoE & $0.3943\,\pm\,0.0000$ & $0.3682\,\pm\,0.0000$ & $\mathbf{0.3559}\,\pm\,0.0000$ \\
\midrule
iTransformer & $0.3233\,\pm\,0.0021$ & $0.3159\,\pm\,0.0018$ & $\mathbf{0.3119}\,\pm\,0.0015$ \\
TQNet & $0.3506\,\pm\,0.0067$ & $0.3417\,\pm\,0.0053$ & $\mathbf{0.3333}\,\pm\,0.0036$ \\
DUET & $0.3640\,\pm\,0.0054$ & $0.3543\,\pm\,0.0046$ & $\mathbf{0.3420}\,\pm\,0.0032$ \\
Moirai & $0.4326\,\pm\,0.0000$ & $0.3766\,\pm\,0.0000$ & $\mathbf{0.3585}\,\pm\,0.0000$ \\
Timer-XL & $0.3177\,\pm\,0.0002$ & $0.3110\,\pm\,0.0001$ & $\mathbf{0.3079}\,\pm\,0.0002$ \\
\midrule
Corrformer & $0.3189\,\pm\,0.0007$ & $0.3138\,\pm\,0.0006$ & $\mathbf{0.3079}\,\pm\,0.0004$ \\
S$^2$Transformer & $0.3158\,\pm\,0.0002$ & $0.3154\,\pm\,0.0002$ & $\mathbf{0.3148}\,\pm\,0.0002$ \\
CDPNet & $0.2985\,\pm\,0.0011$ & $0.2975\,\pm\,0.0013$ & $\mathbf{0.2961}\,\pm\,0.0016$ \\
EasyST & $0.3101\,\pm\,0.0003$ & $0.3091\,\pm\,0.0002$ & $\mathbf{0.3087}\,\pm\,0.0002$ \\
STELLA & $0.4285\,\pm\,0.0534$ & $0.4220\,\pm\,0.0540$ & $\mathbf{0.3931}\,\pm\,0.0384$ \\
\bottomrule
\end{tabular*}
\endgroup
\end{table}

\subsection{Sensitivity to Residual Fitting Sources}
\label{app:residual_source_sensitivity}
On France, we compare three residual fitting sources for PatchTST, iTransformer, and S$^2$Transformer over seeds 2024--2026. All variants share the same frozen backbone and adapter-validation set, with calibration and the corresponding interaction branches retained. \textit{In-sample full} uses all 20,916 backbone-training forecast origins; \textit{In-sample matched} uses its final contiguous 5,140 origins; and \textit{Held-out} uses 5,140 origins from a subsequent block unseen during backbone training. Windows remain within disjoint temporal blocks, and checkpoint selection and adapter tuning use separate validation blocks. The original test period is unchanged; comparisons use the shared Base under this controlled split.

\begin{table}[!htbp]
\centering
\caption{Sensitivity to residual fitting sources on France. MSE is reported as mean $\pm$ sample standard deviation over three seeds. $\Delta$ is the mean per-seed percentage change relative to the shared Base; negative values indicate improvement. Bold marks the lowest mean MSE per backbone.}
\label{tab:residual_source_sensitivity}
\begingroup
\small
\setlength{\tabcolsep}{6pt}
\renewcommand{\arraystretch}{1.1}
\begin{tabular*}{\linewidth}{@{\extracolsep{\fill}}llrr@{}}
\toprule
Backbone & Residual source & MSE $\downarrow$ & $\Delta$ (\%) \\
\midrule
PatchTST & Base & $0.4030\pm0.0001$ & --- \\
 & In-sample full & $\mathbf{0.3703}\pm0.0001$ & $-8.11$ \\
 & In-sample matched & $0.3790\pm0.0001$ & $-5.94$ \\
 & Held-out & $0.3716\pm0.0001$ & $-7.79$ \\
\midrule
iTransformer & Base & $0.3813\pm0.0005$ & --- \\
 & In-sample full & $\mathbf{0.3525}\pm0.0003$ & $-7.54$ \\
 & In-sample matched & $0.3554\pm0.0002$ & $-6.79$ \\
 & Held-out & $0.3526\pm0.0003$ & $-7.51$ \\
\midrule
S$^2$Transformer & Base & $0.3911\pm0.0019$ & --- \\
 & In-sample full & $\mathbf{0.3761}\pm0.0016$ & $-3.82$ \\
 & In-sample matched & $0.3840\pm0.0017$ & $-1.80$ \\
 & Held-out & $0.3786\pm0.0015$ & $-3.19$ \\
\bottomrule
\end{tabular*}
\endgroup
\end{table}

Table~\ref{tab:residual_source_sensitivity} shows that held-out fitting lowers MSE by 7.79\%, 7.51\%, and 3.19\%, respectively, with improvements in every seed. It also outperforms equal-sized in-sample matched fitting for all three backbones. In-sample full is slightly better but uses 4.07 times as many origins. Thus, adapter gains do not require reusing backbone-training residuals, although temporal distribution differences may also affect the comparison.

\subsection{Computational Efficiency}
\label{app:efficiency}
\myparagraph{Measurement protocol.}
We profile seven backbones on France and Europe with seed 2024, batch size 2, and no data-loader workers. Backbone timing measures one training epoch, excluding validation and testing. Adapter timing covers frozen-backbone prediction, training/validation feature accumulation, and ridge fitting with both interaction branches enabled for all six non-MSMV backbones; it excludes prior backbone training and final testing. Comparisons are made within each dataset because batch counts differ. Memory is peak GPU allocation for the complete procedure, including the backbone.

\begin{table}[!htbp]
\centering
\caption{Backbone training cost at batch size 2. Time is minutes per training epoch; memory is peak GPU allocation in GiB.}
\label{tab:backbone_efficiency}
\begingroup
\small
\setlength{\tabcolsep}{5pt}
\renewcommand{\arraystretch}{1.12}
\begin{tabular*}{\linewidth}{@{\extracolsep{\fill}}llrrrr@{}}
\toprule
 & & \multicolumn{2}{c}{\textbf{France}} & \multicolumn{2}{c}{\textbf{Europe}} \\
\cmidrule(lr){3-4}\cmidrule(l){5-6}
Paradigm & Backbone & Time (min) & Memory (GiB) & Time (min) & Memory (GiB) \\
\midrule
SSSV & xPatch & 11.68 & 0.05 & 3.02 & 0.23 \\
     & PatchTST & 9.95 & 1.37 & 1.96 & 7.56 \\
\midrule
SSMV & iTransformer & 6.48 & 0.77 & 2.83 & 4.18 \\
     & DUET & 13.81 & 0.71 & 3.15 & 3.56 \\
\midrule
MSSV & Corrformer & 40.67 & 26.31 & 30.93 & 30.03 \\
     & CDPNet & 141.60 & 5.68 & 24.38 & 2.27 \\
\midrule
MSMV & HiSTGNN & 118.71 & 2.22 & 99.80 & 4.17 \\
\bottomrule
\end{tabular*}
\endgroup
\end{table}

\myparagraph{Cost of interaction modeling.}
Table~\ref{tab:backbone_efficiency} shows that the evaluated MSSV models take substantially longer per epoch than the SSMV models on both datasets, suggesting higher computational costs for station interactions in these implementations. HiSTGNN, the sole native MSMV baseline, is the slowest on Europe (99.80 minutes) and the second slowest on France (118.71 minutes, after CDPNet). Joint modeling can thus incur substantial training cost, although runtime also depends on the backbone architecture.

\begin{table}[!htbp]
\centering
\caption{Complete fitting cost of the full dual-branch adapter with a frozen backbone. $T_{\mathrm{fit}}/T_{\mathrm{epoch}}$ compares total adapter fitting time with one training epoch of the same backbone.}
\label{tab:adapter_efficiency}
\begingroup
\small
\setlength{\tabcolsep}{3pt}
\renewcommand{\arraystretch}{1.12}
\begin{tabular*}{\linewidth}{@{\extracolsep{\fill}}lrrrrrr@{}}
\toprule
 & \multicolumn{3}{c}{\textbf{France}} & \multicolumn{3}{c}{\textbf{Europe}} \\
\cmidrule(lr){2-4}\cmidrule(l){5-7}
Frozen backbone & \makecell{Fit time\\(min)} & \makecell{Memory\\(GiB)} & \makecell{Time ratio\\(\%)} & \makecell{Fit time\\(min)} & \makecell{Memory\\(GiB)} & \makecell{Time ratio\\(\%)} \\
\midrule
xPatch & 2.77 & 0.023 & 23.7 & 0.64 & 0.093 & 21.1 \\
PatchTST & 2.33 & 0.880 & 23.5 & 0.80 & 4.550 & 40.6 \\
iTransformer & 2.93 & 0.700 & 45.3 & 0.62 & 2.560 & 22.0 \\
DUET & 3.32 & 0.690 & 24.1 & 0.71 & 2.510 & 22.6 \\
Corrformer & 16.01 & 28.720 & 39.4 & 12.16 & 21.350 & 39.3 \\
CDPNet & 34.48 & 5.680 & 24.3 & 5.98 & 2.230 & 24.5 \\
\bottomrule
\end{tabular*}
\endgroup
\end{table}

\myparagraph{Adapter fitting cost.}
As shown in Table~\ref{tab:adapter_efficiency}, complete fitting takes only 21.1--45.3\% of one backbone training epoch across all 12 cases, even including frozen-backbone prediction and validation accumulation. This supports the efficiency of the closed-form procedure without iterative backpropagation. Peak memory remains backbone-dependent: it is lower than training in 10 of 12 cases, unchanged at the reported precision for CDPNet on France, and higher for Corrformer on France (28.72 versus 26.31 GiB). Thus, the small adapter parameterization (Appendix~\ref{app:adapter_fitting}) offers consistently low fitting time and usually lower memory requirements, while the total memory footprint also reflects backbone inference and feature processing.

\subsection{Sensitivity to Look-Back Window Length}
\label{app:lookback}
\begin{figure}[!htbp]
\centering
\includegraphics[width=\linewidth]{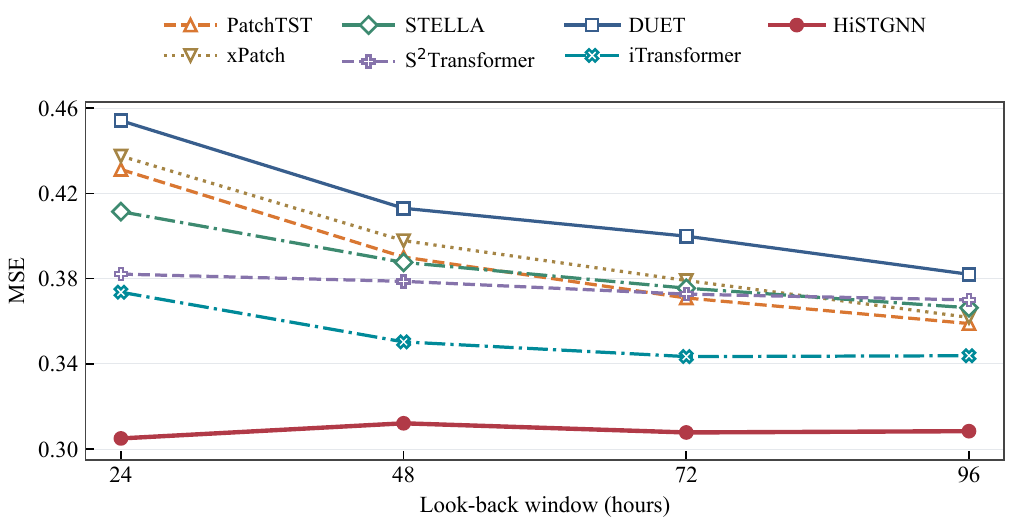}
\caption{MSE versus look-back window on France with a fixed 72-hour forecast horizon. Each curve reports the mean MSE across repeated runs for a native backbone without an adapter. HiSTGNN maintains the lowest MSE across all four windows.}
\label{fig:lookback}
\end{figure}

Figure~\ref{fig:lookback} compares seven native backbones on France across look-back windows of 24, 48, 72, and 96 hours, with a fixed 72-hour forecast horizon. Results are averaged across repeated runs and equally over the four target variables.

Longer histories improve most backbones: PatchTST, xPatch, STELLA, S$^2$Transformer, and DUET perform best at 96 hours, whereas iTransformer reaches its minimum at 72 hours. Nevertheless, HiSTGNN consistently achieves the lowest MSE, varying only from 0.3050 to 0.3121. Its advantage therefore persists beyond the default 48-hour setting, supporting the value of joint station--variable modeling across the evaluated look-back windows. The relative ordering of the other paradigms varies with window length, indicating that sensitivity to historical context remains model-dependent.

\subsection{Sensitivity to KNN Neighborhood Size}
\label{app:knn_sensitivity}
\begin{figure}[!htbp]
\centering
\includegraphics[width=\linewidth]{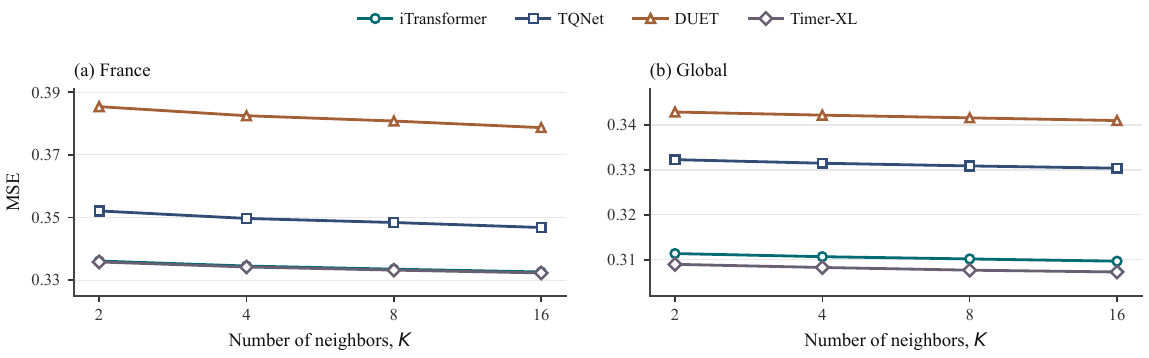}
\caption{Sensitivity of the station-interaction adapter to the KNN neighborhood size $K$. Each curve shows the three-run mean MSE averaged over four variables; lower is better.}
\label{fig:knn_sensitivity}
\end{figure}

We vary the number of neighbors $K\in\{2,4,8,16\}$ in the station-interaction adapter for iTransformer, TQNet, DUET, and Timer-XL on France and Global. The backbone remains frozen, and the ridge coefficient is selected on the validation set for each configuration. Figure~\ref{fig:knn_sensitivity} reports test MSE averaged over three runs and the four target variables.

MSE decreases as $K$ increases for all four backbones on both datasets, with smaller changes from $K=8$ to $16$ than from $K=2$ to $8$ in most cases. Every tested configuration improves upon its corresponding frozen backbone. This indicates that the benefit of station interaction is stable across the examined neighborhood sizes. Considering both efficiency and performance, we use $K=8$ in the main experiments.

\subsection{Prediction Visualizations}
\label{app:prediction_visualization}
\begin{figure}[!htbp]
\centering
\includegraphics[width=\linewidth,height=0.78\textheight,keepaspectratio]{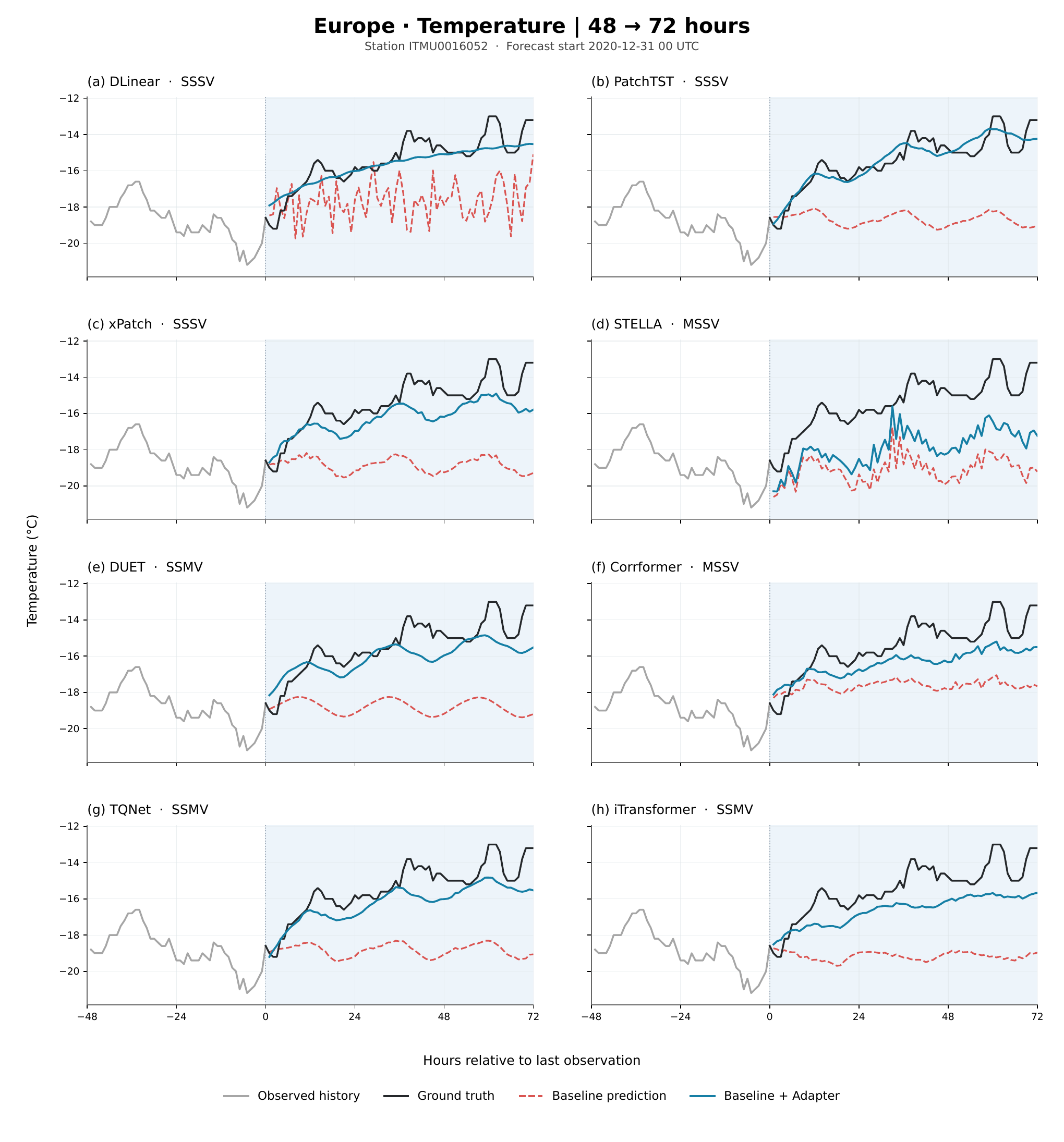}
\caption{\textbf{Air temperature forecasts on Europe.} Station ITMU0016052; forecast start 2020-12-31 00:00 UTC. Each panel compares a native backbone with its full adapter over the same 72-hour window following 48 hours of history. Adapted forecasts better follow the sustained warming trend.}
\label{fig:prediction_t}
\end{figure}

\begin{figure}[!htbp]
\centering
\includegraphics[width=\linewidth,height=0.78\textheight,keepaspectratio]{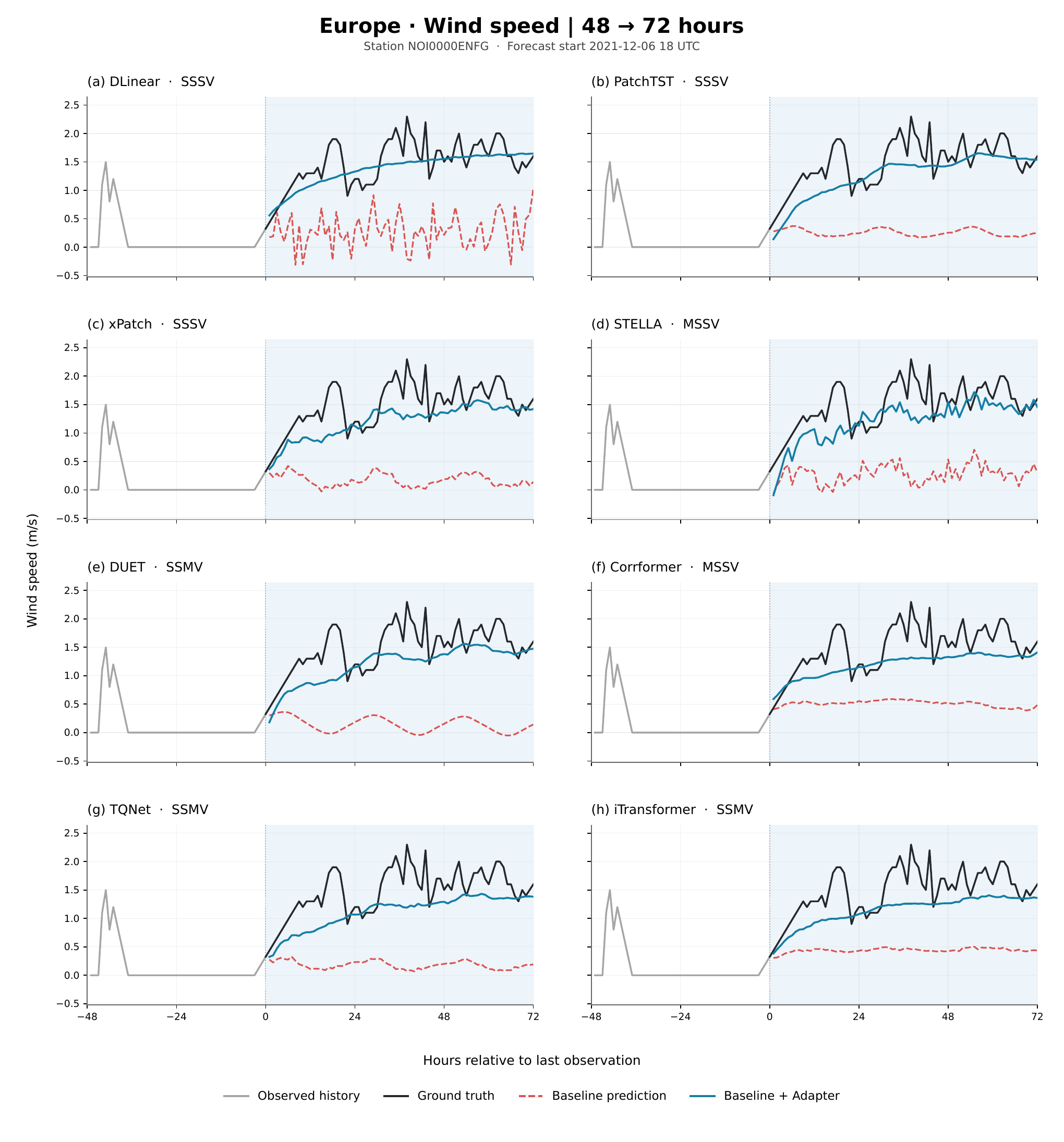}
\caption{\textbf{Wind speed forecasts on Europe.} Station NOI0000ENFG; forecast start 2021-12-06 18:00 UTC. Each panel compares a native backbone with its full adapter over the same 72-hour window following 48 hours of history. Adapted forecasts capture the increase in wind speed, while rapid fluctuations remain difficult to reproduce.}
\label{fig:prediction_ws}
\end{figure}

\begin{figure}[!htbp]
\centering
\includegraphics[width=\linewidth,height=0.78\textheight,keepaspectratio]{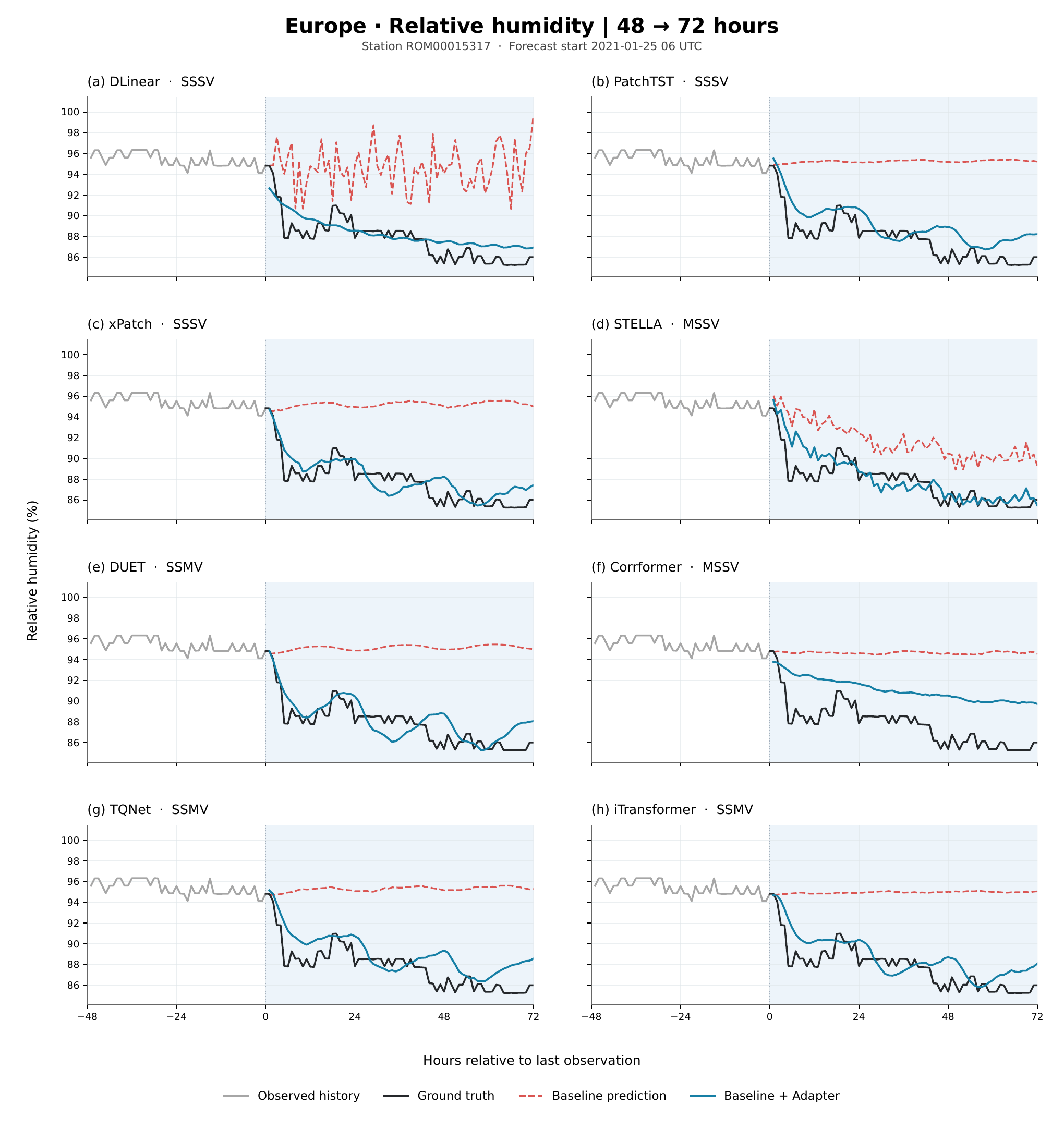}
\caption{\textbf{Relative humidity forecasts on Europe.} Station ROM00015317; forecast start 2021-01-25 06:00 UTC. Each panel compares a native backbone with its full adapter over the same 72-hour window following 48 hours of history. Adapted forecasts better track the humidity decline, with residual differences in timing and magnitude.}
\label{fig:prediction_rh}
\end{figure}

\begin{figure}[!htbp]
\centering
\includegraphics[width=\linewidth,height=0.78\textheight,keepaspectratio]{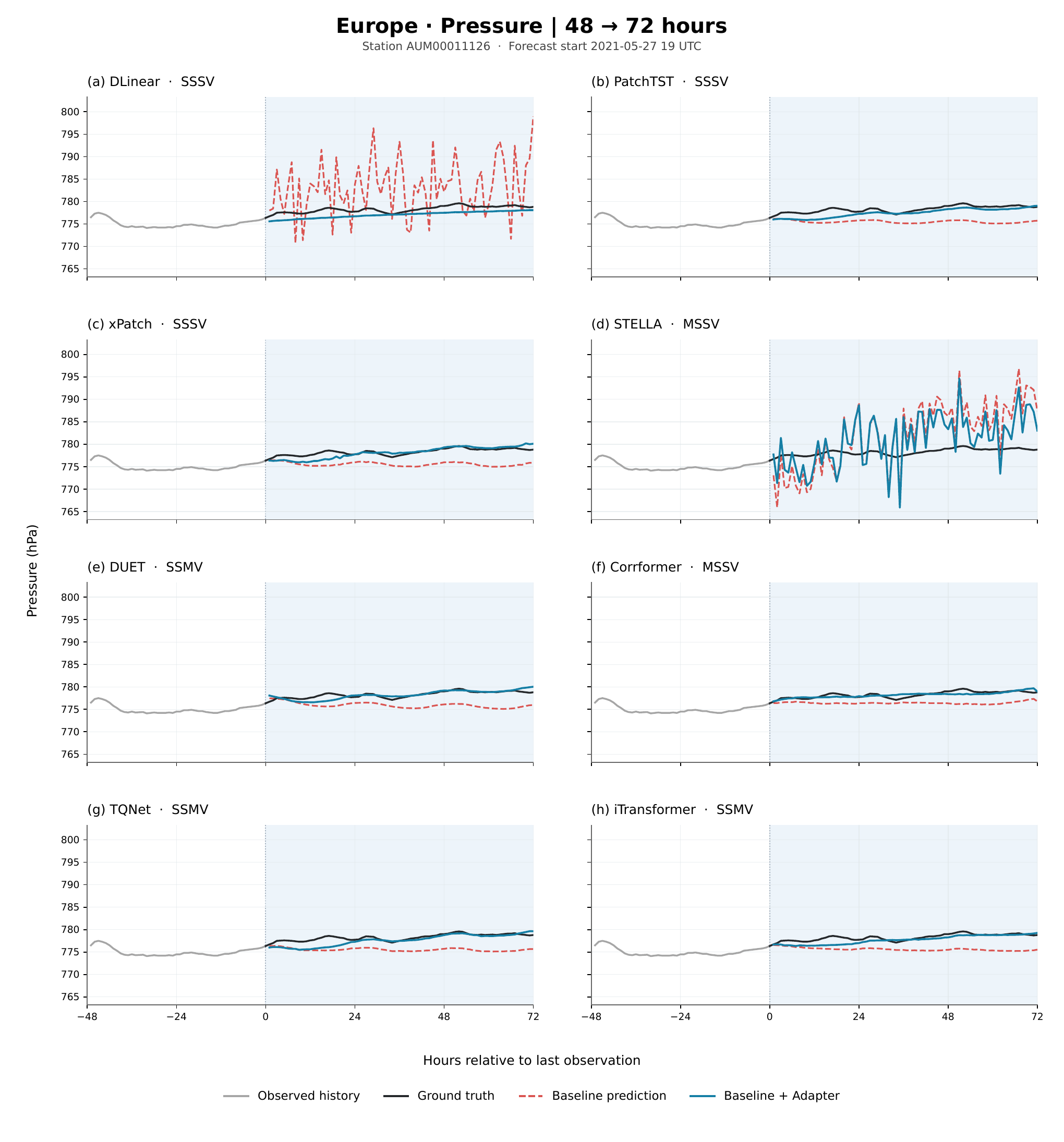}
\caption{\textbf{Station pressure forecasts on Europe.} Station AUM00011126; forecast start 2021-05-27 19:00 UTC. Each panel compares a native backbone with its full adapter over the same 72-hour window following 48 hours of history. Most adapted forecasts follow the observed pressure level and gradual variation more closely; STELLA retains pronounced fluctuations.}
\label{fig:prediction_p}
\end{figure}

Figures~\ref{fig:prediction_t}--\ref{fig:prediction_p} compare native and adapted forecasts for four example windows on Europe, one per target variable. Each figure presents the same station and forecast window for eight backbones: DLinear, PatchTST, xPatch, STELLA, DUET, Corrformer, TQNet, and iTransformer. The plots show 48 hours of observed history followed by a 72-hour forecast in physical units. Gray denotes historical observations, black the future ground truth, red dashed lines the native forecasts, and blue lines the forecasts with the full adapter; the shaded region marks the forecast horizon.

The temperature and humidity examples illustrate improved tracking of a sustained warming trend and a humidity decline that several native forecasts largely miss. In the wind-speed example, adapted forecasts better capture the increase after a prolonged calm period, although short-lived peaks remain smoothed. For pressure, most adapted trajectories follow the slowly varying observations more closely, while STELLA retains substantial fluctuations. These examples illustrate how output correction can improve forecast levels and temporal trends across different backbones, alongside the aggregate results. Together, they provide qualitative evidence for the effectiveness of the adapter across weather variables and backbone architectures.

\end{document}